%% file: main.tex
\pdfoutput=1
\documentclass[acmtog]{acmart}
\AtEndPreamble{\hypersetup{colorlinks=true, linkcolor=ACMPurple, citecolor=ACMPurple, urlcolor=ACMDarkBlue}}
\makeatletter
\AtEndPreamble{\let\ACM@orig@copyrightpermission\@copyrightpermission
  \def\@copyrightpermission{\begingroup\hypersetup{urlcolor=black}\ACM@orig@copyrightpermission\endgroup}}
\makeatother

\usepackage{enumitem}
\newcommand{\topic}[1]
{
\noindent\textbf{#1}
}

\input{header}
\usepackage{tcolorbox}
\Crefname{figure}{Fig.}{Figs.}

\author{Chuhao Chen}
\affiliation{%
  \institution{University of Pennsylvania}
  \city{Philadelphia}
  \country{USA}}
\email{morphling233@gmail.com}

\author{Peter Wonka}
\affiliation{%
  \institution{Snap Inc.}
  \city{Santa Monica}
  \country{USA}}
\affiliation{%
  \institution{KAUST}
  \city{Thuwal}
  \country{Saudi Arabia}}
\email{pwonka@gmail.com}

\author{Chaoyang Wang}
\affiliation{%
  \institution{Snap Inc.}
  \city{Santa Monica}
  \country{USA}}
\email{gordon.w.1991@gmail.com}

\author{Chen Wang}
\affiliation{%
  \institution{University of Pennsylvania}
  \city{Philadelphia}
  \country{USA}}
\email{chenw30@seas.upenn.edu}

\author{Qiao Feng}
\affiliation{%
  \institution{University of Pennsylvania}
  \city{Philadelphia}
  \country{USA}}
\email{fengqiao@seas.upenn.edu}

\author{Sergey Tulyakov}
\affiliation{%
  \institution{Snap Inc.}
  \city{Santa Monica}
  \country{USA}}
\email{stulyakov@snap.com}

\author{Lingjie Liu}
\affiliation{%
  \institution{University of Pennsylvania}
  \city{Philadelphia}
  \country{USA}}
\email{lingjie.liu@seas.upenn.edu}

\copyrightyear{2026}
\acmYear{2026}
\setcopyright{cc}
\setcctype{by}
\acmConference[SA Conference Papers '26]{SIGGRAPH Asia 2026 Conference Papers}{December 01--04, 2026}{Kuala Lumpur, Malaysia}
\acmBooktitle{SIGGRAPH Asia 2026 Conference Papers (SA Conference Papers '26), December 01--04, 2026, Kuala Lumpur, Malaysia}
\acmDOI{10.1145/3829340.3842176}
\acmISBN{979-8-4007-2842-6/2026/12}

\begin{document}

\title{\name: Streaming Physics-Grounded Video Generation with Structured Scene Memory and Fine-Grained Motion Control}

\input{sections/00_abstract}



\begin{teaserfigure}
    \includegraphics[width=\textwidth]{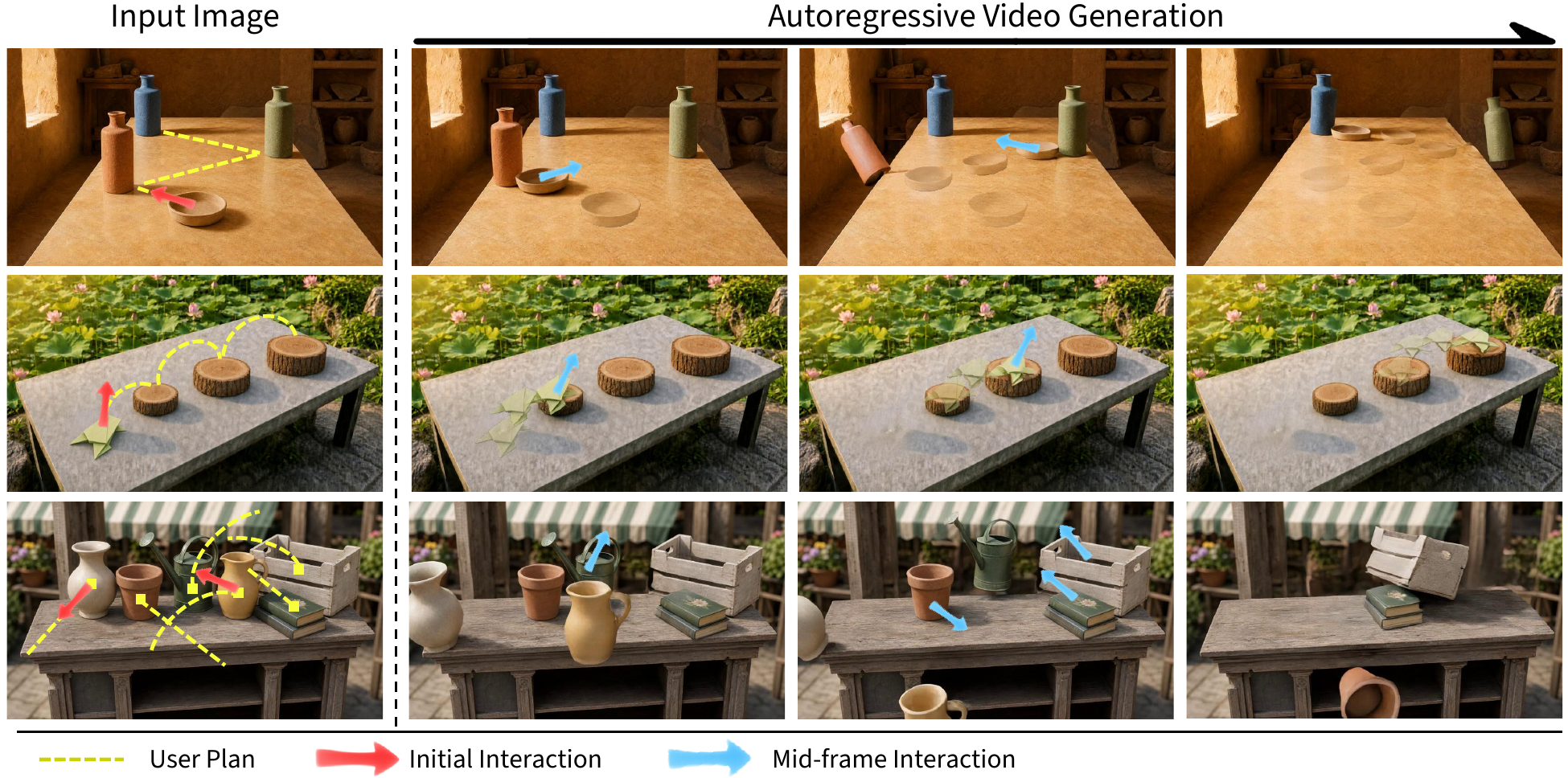}
    \caption{\name generates physics-grounded videos from a single image through sparse, interactive, scene-level velocity control: users specify per-object velocity directions at chosen timesteps, and the model autoregressively produces physically plausible multi-object dynamics.
    \textit{Top:} a ceramic dish zig-zags across a tabletop, precisely striking and toppling vases near the edge.
    \textit{Middle:} an origami frog leaps onto three successive wooden stumps on a stone table.
    \textit{Bottom:} assorted objects at a market stall are swept off the table one or several at a time.}
    \label{fig:teaser}
\end{teaserfigure}
\maketitle
\renewcommand{\shortauthors}{Chen et al.}

\input{sections/01_introduction}
\input{sections/02_related_work}
\input{sections/03_method}
\input{sections/04_experiments}
\input{sections/05_conclusion}

\begin{acks}
This work was funded in part by a gift from Snap Inc.
We thank the participants of our user study for their time, and the anonymous reviewers for their constructive feedback.
\end{acks}

\bibliographystyle{ACM-Reference-Format}
\bibliography{main}

\appendix
\makeatletter
\twocolumn[{\@titlefont Supplementary Material: \name\par}\bigskip]
\makeatother
\input{appendix}
\end{document}

%% file: header.tex
\usepackage{xspace}
\usepackage{colortbl}
\definecolor{YellowOrange}{RGB}{255,174,66}
\definecolor{LightYellow1}{RGB}{255,255,224}
\definecolor{Goldenrod}{RGB}{255,223,124}
\usepackage{multirow}
\usepackage{amsmath}
\usepackage[capitalize]{cleveref}

\graphicspath{{figures/}{./}}

\newcommand{\name}{PhysStream\xspace}

%% file: sections/00_abstract.tex
\begin{abstract}
    Interactive control for video generation is moving from coarse prompts toward fine-grained, physically meaningful manipulation of dynamic scenes.
    Yet existing controllable methods either require the full control schedule before generation starts, or use pixel-space signals that dictate object positions rather than physical dynamics.
    To address these limitations, we propose \name, an autoregressive model for physics-grounded image-to-video synthesis that incorporates structured scene memory---positional maps and object tracking maps derived online from previously generated frames---and supports fine-grained motion control via sparse velocity-increment signals that encode physical quantities, letting the model learn the underlying dynamics.
    We train our model in two stages: a bidirectional model is first finetuned with motion-control conditioning, then a causal autoregressive model is trained with additional structured scene memory, further improving physical consistency.
    \name enables interactive, mid-generation control over multi-object tabletop rigid-body scenes---a capability not supported by prior methods---reducing motion distribution distance (FVMD) by 33\% and trajectory error by 12\% over the strongest baselines on synthetic benchmarks, and is preferred by human evaluators in over 85\% of in-the-wild comparisons.
        Please check our website for more details: {\hypersetup{urlcolor=magenta}\url{https://czzzzh.github.io/PhysStream}}.
    \end{abstract}

%% file: sections/01_introduction.tex
\section{Introduction} 
\label{sec:intro}

Video diffusion models~\cite{wan2025wan,yang2024cogvideox,ho2022video,blattmann2023stable} have emerged as powerful tools for high-fidelity video synthesis, with applications spanning simulation, robotics, and creative content generation.
Building on these advances, controllable video generation leverages additional conditions---depth maps~\cite{Zhang2023ControlVideoTC,wang2023videocomposer}, camera trajectories~\cite{bahmani2025ac3d,he2024cameractrl,he2025cameractrl}, object tracks or keypoints~\cite{gu2025diffusion,li2026flashmotion,zhang2025tora,niu2024mofa, namekata2024sg}, and physical interactions such as forces or velocities~\cite{wang2025physctrl,gillman2025force,gillman2026goal,romero2025learning} to steer the generated videos towards the given condition.
These methods have achieved impressive results for manipulating foreground objects or camera movement, yet they predominantly operate in a non-autoregressive manner: the full control schedule must be specified before generation begins, and the entire clip is synthesized in one pass.
This design precludes truly interactive use cases in which a user observes previously generated frames and decides the next intervention on the fly.

Recent advances in autoregressive video diffusion~\cite{chen2024diffusion,huang2025self,zhu2026causal,liu2025rolling,li2026rolling} have enabled incremental, frame-by-frame generation that opens the door to interactive controllable video synthesis.
Building on this progress, we identify four key properties for controllable video generation that simultaneously serve interactive creative workflows and physics-grounded simulation:
\textbf{(1)~Sparse control}---the signal should be easy for a user to construct (e.g., a drag trajectory or a velocity vector on an object), rather than a dense per-pixel map such as depth or optical flow;
\textbf{(2)~Physics-grounded}---the signal should encode a physical quantity (force, velocity) that lets the model learn the underlying dynamics, rather than directly dictating object positions along a prescribed path;
\textbf{(3)~Interactive}---generation should proceed frame-by-frame so users can observe partial results and intervene on the fly; we use the term in this control sense and do not require real-time throughput;
\textbf{(4)~Scene-level}---control should target individual objects within a multi-object scene.
\Cref{tab:desiderata} compares a selection of representative methods along these axes.
Among them, only the concurrent work RealWonder~\cite{liu2026realwonder} approaches all four; however, its interaction is mediated by an external 3D reconstruction and physics simulator whose scene state may diverge from the actual generated video---for instance, object positions in the reconstructed scene can drift from those in the synthesized frames, and unmodeled background objects cannot participate in physical interactions.

\input{tables/desiderata}

To satisfy all four properties through direct interaction with the generated video, we propose \name, an autoregressive image-to-video model.
At each autoregressive step, \name conditions on (i)~sparse velocity-increment maps that let the user apply localized interactions to selected objects, and (ii)~a structured scene memory comprising positional maps (from monocular depth estimation) and object-tracking maps (from instance segmentation and tracking), both derived from previously generated frames and updated online after each generated frame.
Adapting a pretrained bidirectional video model to this formulation involves three distribution shifts: the velocity-increment control, the structured scene memory, and the change from bidirectional to causal attention. They cannot all be learned at once: the scene memory records the full object history, which may lead the model to partly ignore the historical velocity signals, and it cannot be learned under bidirectional attention at all, since per-frame memory maps would leak future scene state. We therefore train in two stages: a bidirectional backbone first learns the velocity-increment control alone, and a causal autoregressive model is then trained on top of it, learning the scene memory and causal attention jointly---a recipe that keeps each transition small without multiplying training stages.

We conduct extensive experiments and demonstrate great improvements in both motion-control adherence and physical plausibility. Our main contributions are:

\begin{itemize}[nosep]
    \item We propose \name, the first method that enables direct, end-to-end, scene-level physics-grounded interactive video control in multi-object tabletop rigid-body scenes, where the user's physical input and the model's scene memory both operate on the generated video itself.
    \item We introduce structured scene memory---positional maps and object-tracking maps updated online from previously generated frames---as a novel conditioning mechanism for autoregressive video generation, and show that it effectively improves geometric consistency and physical plausibility.
    \item We curate a dataset of 100k synthetic indoor scene videos with complex multi-object rigid-body motion, collisions, and multi-frame velocity perturbations, aiming to further improve the physical correctness of video generation models.
\end{itemize} 

%% file: tables/desiderata.tex
\begin{table}[t]
    \caption{Representative controllable video generation methods~\cite{Zhang2023ControlVideoTC,burgert2025go,gu2025diffusion,bahmani2025ac3d,he2025cameractrl,wu2024draganything,zhou2025streaming,zhang2025tora,shin2025motionstream,li2026flashmotion,niu2024mofa,yang2025longlive,he2025matrix,gillman2025force,wang2025physctrl,liu2026realwonder,romero2025learning} compared along the four properties.
    \textbf{Sparse}: easy-to-construct signal (not dense per-pixel); \textbf{Phys.}: physics-grounded; \textbf{Inter.}: interactive; \textbf{Scene}: scene-level.
    $^{*}$RealWonder supports interaction and scene-level control through an intermediate 3D reconstruction and physics simulator, whose scene state may diverge from the generated video.}
    \label{tab:desiderata}
    \centering
    \small 
    \setlength{\tabcolsep}{4pt}
    \begin{tabular}{llcccc}
        \toprule
        Method & Control & Sparse & Phys. & Inter. & Scene \\
        \midrule
        ControlVideo~             & depth       &            &            &            & \checkmark \\
        Go-with-the-Flow~         & flow        &            &            &            & \checkmark \\
        DaS~                      & dense track &            &            &            & \checkmark \\
        AC3D~                     & camera      & \checkmark &            &            &            \\
        CameraCtrl II~            & camera      & \checkmark &            & \checkmark &            \\
        DragAnything~             & drag        & \checkmark &            &            & \checkmark \\
        DragStream                & drag        & \checkmark &            & \checkmark & \checkmark \\
        Tora                      & trajectory  & \checkmark &            &            & \checkmark \\
        MotionStream              & trajectory  & \checkmark &            & \checkmark & \checkmark \\
        FlashMotion               & mask track  & \checkmark &            &            & \checkmark \\
        MOFA-Video~               & keypoint    & \checkmark &            &            & \checkmark \\
        LongLIVE~                 & text prompt & \checkmark &            & \checkmark & \checkmark \\
        Matrix-Game 2.0~          & game action & \checkmark &            & \checkmark &            \\
        Force Prompting           & force       & \checkmark & \checkmark &            &            \\
        PhysCtrl                  & force       & \checkmark & \checkmark &            &            \\
        RealWonder$^{*}$          & force       & \checkmark & \checkmark & \checkmark$^{*}$ & \checkmark$^{*}$ \\
        KineMask~                 & velocity    & \checkmark & \checkmark &            & \checkmark \\
        \name (Ours)        & velocity    & \checkmark & \checkmark & \checkmark & \checkmark \\
        \bottomrule
    \end{tabular} 
\end{table}

%% file: sections/02_related_work.tex
\section{Related Work}
\label{sec:related}

\noindent\textbf{Controllable Video Generation}
Controllable video generation conditions video models using auxiliary signals beyond text prompts to improve controllability and user intention.
Depth-based methods~\cite{Zhang2023ControlVideoTC,wang2023videocomposer} and camera-trajectory controllers~\cite{bahmani2025ac3d,he2024cameractrl,he2025cameractrl} guide global scene motion, while object-level approaches use drag points~\cite{yin2023dragnuwa,wu2024draganything}, bounding-box tracks~\cite{wang2024boximator,ma2024trailblazer}, mask tracks~\cite{li2025magicmotion,li2026flashmotion}, dense optical flow and point tracks~\cite{burgert2025go,gu2025diffusion,geng2025motion}, or sparse keypoint trajectories~\cite{wang2024motionctrl,niu2024mofa,zhang2025tora,fu20243dtrajmaster,namekata2024sg} to manipulate individual entities.
Most of these methods use ControlNet~\cite{zhang2023adding}, cross-attention injection, or channel-wise concatenation to inject the control signals into a pretrained video model.
While these approaches achieve strong controllability, they require control signals over all timesteps, rather than encoding a physical quantity that lets the model predict \textit{how} objects move.
In contrast, we target interactive, physics-grounded, scene-level control: users provide only a sparse velocity vector at chosen timesteps, and the model learns to produce physically consistent multi-object dynamics from that signal alone.

\noindent\textbf{Physics-Grounded Video Generation}
A growing line of work seeks to improve the physical plausibility of video generative models.
One family of approaches obtains motion signals from physics simulators and injects them into video models, including PhysGen~\cite{liu2024physgen} for rigid body dynamics, PhysGen3D and PhysMotion~\cite{chen2025physgen3d, tan2024physmotion} for deformable bodies, and PhysAnimator~\cite{xie2025physanimator} for cartoon animations.
WonderPlay~\cite{li2025wonderplay}, RealWonder~\cite{liu2026realwonder} and PSIVG~\cite{foo2026physical} study the interplay between physics solver and video diffusion for better visual quality.
However, these methods require calling physical simulators at inference time, which some other works try to avoid.
PhysCtrl~\cite{wang2025physctrl} trains a trajectory predictor given user actions to guide video generation. Force Prompting~\cite{gillman2025force} and Goal Force~\cite{gillman2026goal} also curate action and video pairs from simulation to directly finetune a pretrained video model.
The third family uses geometric consistency as an indirect physics proxy: depth/normal regularization~\cite{zhang2025world,ren2025gen3c} or 3D-aware world models~\cite{zhu2025aether,team2026advancing}.
Our work differs from prior works in that we do not rely on an external simulator or trajectory at inference time, nor do we impose any consistency loss in an implicit manner.
Instead, we explicitly condition on a structured scene memory estimated on-the-fly from the model's own prediction for physics-grounded generation.

\topic{Autoregressive and Streaming Video Generation}
Autoregressive video generation produces frames frame-by-frame or chunk-by-chunk, naturally supporting streaming output and interactive feedback. Teacher-Forcing~\cite{williams1989learning, jin2024pyramidal} and Diffusion Forcing~\cite{chen2024diffusion, song2025history} are well-established paradigms for training autoregressive video diffusion models with clean-context as history. 
More recently, distillation-based approaches have emerged to distill strong pretrained bidirectional models into few-step causal models: CausVid~\cite{yin2025slow} applies distribution matching distillation~\cite{yin2024one} to obtain a few-step causal generator, Self-Forcing~\cite{huang2025self} further introduces training time rollout to bridge the train-inference gap, and Causal-Forcing~\cite{zhu2026causal} finetunes a bidirectional model into a causal architecture to eliminate the architecture gap before distillation. 
Most related to our work, DragStream~\cite{zhou2025streaming} and MotionStream~\cite{shin2025motionstream} concatenate motion-control channels to the autoregressive generator, demonstrating on-the-fly trajectory-based and drag-based interaction during streaming generation.
However, existing autoregressive methods treat each generated frame independently of the scene's physical state: no history-derived geometric or object-tracking signal is fed back to the generator for future generation.
We build on the autoregressive paradigm and introduce structured scene memory as a feedback loop, enabling the model to leverage its generation history to improve physical consistency. 

%% file: sections/03_method.tex
\section{Method}
\label{sec:method}
\subsection{Overview}
\label{sec:method_overview}

\topic{Task Definition}
\label{sec:task_def}
We consider physics-grounded image-to-video (I2V) generation under autoregressive sampling.
A sample consists of an initial frame $x_0 \in \mathbb{R}^{H\times W\times 3}$, a sequence of $N$ subsequent frames $x_{1:N} = (x_1, \dots, x_N)$ to be generated, and an optional text prompt $y$.
A causal model factorizes the joint distribution as
{
\begin{equation}
\!\!p_\theta(x_{1:N}\mid x_0, y) = \prod_{i=1}^{N} p_\theta\!\left(x_i \mid x_0, x_{<i}, y \right),
\label{eq:ar_factor}
\end{equation}
}
where $x_{<i} := (x_1, \dots, x_{i-1})$.

Beyond the standard I2V conditioning, our model accepts two additional history-derived signals.
The first is a \textbf{structured scene memory},
comprising a normalized positional map $c^{\mathrm{pos}\vphantom{\mathrm{pk}}}_t \in [0,1]^{H\times W\times 3}$ that encodes per-pixel 3D camera-frame coordinates, and an object-tracking map $c^{\mathrm{track}\vphantom{\mathrm{pk}}}_t \in [0,1]^{H\times W\times 3}$ where each tracked object is painted with a distinct palette color on a black background.
Both are estimated automatically from previously generated frames.
The second is a \textbf{user-specified velocity-increment map} $c^{\Delta v\vphantom{\mathrm{pk}}}_t \in [0,1]^{H\times W\times 3}$, an object-level 3D velocity signal painted onto the spatial masks of selected objects (see \cref{sec:vel_cond}) that the user may inject at any frame $t$.
All three signals are strictly historical with respect to the frame being synthesized: the conditional distribution becomes
{
\begin{equation}
x_i \;\sim\; p_\theta\!\left(x_i \,\middle|\, x_0, x_{<i}, y,\; c^{\Delta v\vphantom{\mathrm{pk}}}_{<i},\, c^{\mathrm{pos}\vphantom{\mathrm{pk}}}_{<i},\, c^{\mathrm{track}\vphantom{\mathrm{pk}}}_{<i} \right),
\label{eq:ar_control}
\end{equation}
}
where $c_{<i} := (c_0, \dots, c_{i-1})$ collects all past frames for each condition (the user injects each velocity increment before the corresponding frame is generated).

We instantiate this formulation under \textbf{rigid-body dynamics} captured by a \textbf{static camera}, which provides a clean physical setting for studying multi-object scene-level interaction.
To this end, we curate a $100$k-scale synthetic dataset of indoor scenes augmented with rigid-body simulations; see \cref{sec:datasets} for details.

\begin{figure*}[t]
    \centering
    \includegraphics[width=\textwidth]{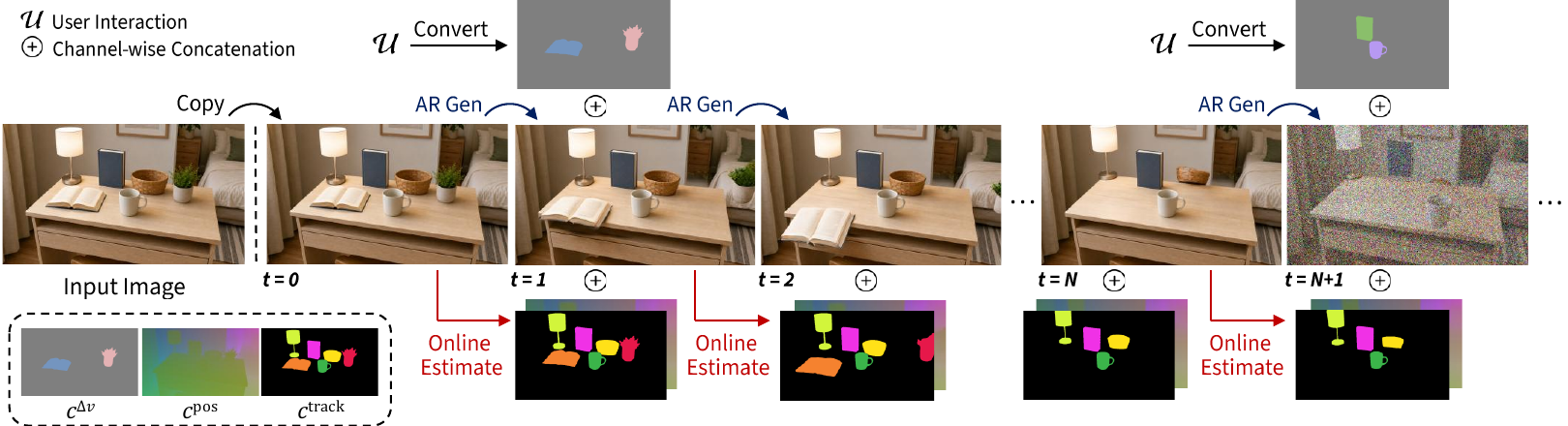}
    \caption{Autoregressive inference pipeline of \name. Given an input image ($t{=}0$), the model autoregressively generates each subsequent latent frame by denoising a noisy latent conditioned on: (1)~the user-specified velocity-increment map $c^{\Delta v\vphantom{\mathrm{pk}}}$ (channel-concatenated with a one-frame temporal shift), and (2)~the structured scene memory $(c^{\mathrm{pos}\vphantom{\mathrm{pk}}}, c^{\mathrm{track}\vphantom{\mathrm{pk}}})$, which is estimated online from the most recently decoded frames via a monocular depth estimator and SAM2. After each latent frame is committed, the decoded RGB frames are fed back to the online estimators to update the scene memory for the next step.}
    \label{fig:overview}
\end{figure*}

\topic{Two-Stage Training}
\label{sec:two_stage} \name is trained in two stages.
\textbf{Stage~1} (\cref{sec:stage1}) finetunes the bidirectional Wan2.2-TI2V-5B~\cite{wan2025wan} video diffusion model to consume only the user-specified velocity-increment condition $c^{\Delta v\vphantom{\mathrm{pk}}}$.
\textbf{Stage~2} (\cref{sec:stage2}) converts this base into a causal autoregressive model in a Teacher-Forcing manner following Causal-Forcing~\cite{zhu2026causal}, generating frames frame-by-frame with KV caching, and additionally introduces the structured scene memory $(c^{\mathrm{pos}\vphantom{\mathrm{pk}}}, c^{\mathrm{track}\vphantom{\mathrm{pk}}})$ estimated online from the model's own previously generated frames.
Across both stages, every condition is injected via channel-wise concatenation of VAE-encoded latents combined with a one-frame temporal shift, which, together with causal attention, guarantees that each noisy latent only sees conditions derived from previous-frame content.
After two-stage training, our autoregressive video generation process is illustrated in \cref{fig:overview}.   

\subsection{Stage 1: Bidirectional Generation with Motion Control}
\label{sec:stage1}
In Stage~1, we model the conditional distribution
{
\begin{equation}
p^{\mathrm{bi}}_\theta\!\left(x_{1:N} \,\middle|\, x_0, y, c^{\Delta v\vphantom{\mathrm{pk}}}_{0:N}\right),
\label{eq:stage1_dist}
\end{equation}
}
where the velocity-increment condition $c^{\Delta v\vphantom{\mathrm{pk}}}_{0:N}$ is the sole user-provided motion signal and the model denoises all frames jointly.

\topic{Velocity-Increment Condition}
\label{sec:vel_cond} Let $\mathcal{O}$ denote the set of dynamic rigid-body objects present in the first frame, and let $M^{(o)} \in \{0,1\}^{H\times W}$ be the binary instance mask of object $o\in\mathcal{O}$ in $x_0$.
This mask is defined once on the first frame and reused for all velocity-increment events throughout the video, regardless of the object's actual position at the time of each event (see \cref{sec:mask_choice} for the rationale).
At training time, $M^{(o)}$ is read from the rendered ground-truth mask; at inference time, the user designates the target object $o$ and $M^{(o)}$ is obtained with the help of an off-the-shelf segmentation model.

We assume that every user-specified velocity change is bounded along each camera axis by a fixed maximum input speed $V_{\max}$, uniform across axes.
The user provides a sparse set of velocity-increment events
{
\begin{equation}
\mathcal{U} \;=\; \bigl\{ (t_j,\, o_j,\, \Delta\mathbf{v}_j) \bigr\}_{j=1}^{J},
\label{eq:user_events}
\end{equation}
}
where $t_j \in \{0, \dots, N\}$, $o_j \in \mathcal{O}$, and $\Delta\mathbf{v}_j \in [-V_{\max}, V_{\max}]^3$ is the camera-frame velocity change applied uniformly across the rigid body of object $o_j$ at frame $t_j$.
Each event is linearly mapped to a normalized value $\tilde{\mathbf{v}}_j \in [0,1]^3$, where $\tfrac{1}{2}\mathbf{1}$ encodes zero velocity change and the extremes $0$ and $1$ correspond to $-V_{\max}$ and $+V_{\max}$ respectively.

The per-frame velocity-increment map $c^{\Delta v\vphantom{\mathrm{pk}}}_t \in [0,1]^{H\times W \times 3}$ is then obtained by painting each event onto the corresponding object's first-frame mask $M^{(o_j)}$, leaving all remaining pixels at the neutral value:
{
\begin{equation}
c^{\Delta v\vphantom{\mathrm{pk}}}_t(p) \;=\; \tilde{\mathbf{v}}_j \;\;\text{if } \exists\, j:\, t_j{=}t,\, M^{(o_j)}(p){=}1; \;\;\text{else } \tfrac{1}{2}\mathbf{1}.
\label{eq:vel_map}
\end{equation}
}

\topic{First-Frame Mask vs.\ Per-Frame Mask}
\label{sec:mask_choice}
As shown in \cref{fig:overview}, we always anchor velocity-increment events to the object's position in the first frame given by mask $M^{(o)}$: even when an object has moved away from its initial position by frame $t_j$, the velocity signal is painted at the first-frame location, not the current one.
Note that this is purely a training-time convention; at inference time, the user can still visually select the object at its current position in the generated video, and the system internally maps the interaction back to the first-frame mask.
A natural alternative to this design is to paint each event on the object's mask at frame $t_j$.
While this signal is in principle more accurate, we find that under bidirectional training, it leaks the moving object's spatial trajectory into the condition channel.
This leakage is particularly harmful when transitioning from bidirectional to causal training in Stage~2: the causal model can no longer access future-frame masks, so the condition distribution shifts abruptly, widening the gap between the two stages and degrading generation quality.
Anchoring every event to the frame-$0$ mask removes this leakage path and keeps the condition distribution consistent across both stages.
For the same reason, we exclude the structured scene memory $(c^{\mathrm{pos}\vphantom{\mathrm{pk}}}, c^{\mathrm{track}\vphantom{\mathrm{pk}}})$ from Stage~1: per-frame positional and tracking maps would similarly leak the future scene state under bidirectional attention.
The structured scene memory is introduced only in Stage~2, where causal masking together with the temporal shift in \cref{sec:condition_injection} prevents any future leakage.
See \cref{sec:supp_mask} for more experimental evidence.

\subsection{Stage 2: Autoregressive Generation with Structured Scene Memory}
\label{sec:stage2}

Stage~2 directly realizes \cref{eq:ar_control} in causal autoregressive form: each frame $x_i$ is sampled given the history $(x_0, x_{<i}, y)$ together with the three signals $c^{\Delta v\vphantom{\mathrm{pk}}}_{<i}$, $c^{\mathrm{pos}\vphantom{\mathrm{pk}}}_{<i}$, $c^{\mathrm{track}\vphantom{\mathrm{pk}}}_{<i}$.
The motion-control condition $c^{\Delta v\vphantom{\mathrm{pk}}}$ retains the form of \cref{eq:vel_map}; the two scene-memory conditions are not user-supplied but produced \emph{online} by two estimators that operate on the model's previously generated frames.

\topic{Normalized Positional Map}
\label{sec:pos_map}
We adopt a normalized positional map similar to the one used in~\cite{zhang2025world}.
The estimator $\Phi_{\mathrm{pos}}$ runs Depth-Anything-3~\cite{lin2025depth} on the most recent $L$ pixel frames to obtain per-frame metric depth $\hat{D}_t$ and intrinsics $K_t$ (we find $L{=}4$, i.e., one latent frame, sufficient in practice). Each pixel $p = (u, v)$ is back-projected into a 3D camera-frame coordinate
{
\begin{equation}
\mathbf{P}_t(p) \;=\; \hat{D}_t(p)\, K_t^{-1}\, [u, v, 1]^{\top} \;\in\; \mathbb{R}^3,
\label{eq:da3_unproj}
\end{equation}
}
matching the camera-space convention of our training-data rendering (\cref{sec:datasets}).
The coordinates are then centered and uniformly normalized into $[0,1]^3$ using a normalization anchor computed once from the first frame: we define the per-axis extremes $\mathbf{P}_{\min}, \mathbf{P}_{\max} \in \mathbb{R}^3$ over all pixels in frame $0$, and a uniform scale factor
{
\begin{equation}
\rho \;=\; \tfrac{1}{2}\,\max_{a \in \{x,y,z\}} (P_{\max,a} - P_{\min,a}),
\label{eq:pos_scale}
\end{equation}
}
which preserves the isotropic aspect ratio across all three axes.
The normalized positional map is then
{
\begin{equation}
c^{\mathrm{pos}\vphantom{\mathrm{pk}}}_t(p) \;=\; \frac{\mathbf{P}_t(p) - \tfrac{1}{2}(\mathbf{P}_{\min} + \mathbf{P}_{\max})}{2\rho} + \tfrac{1}{2}\,\mathbf{1} \;\in\; [0,1]^3.
\label{eq:pos_norm}
\end{equation}
}
Under our static-camera setting the depth range remains close to that of the first frame, so this anchor stays stable throughout generation. After obtaining $L$ positional maps, we only append those for newly decoded frames to the condition sequence.
Our design ensures the preservation of the KV cache (i.e., committed positional maps remain unchanged) while maintaining temporal consistency as much as possible. More experimental evidence is provided in \cref{sec:supp_posmap}.

\topic{Object-Tracking Map}
\label{sec:track_map}
Given decoded frames together with the first-frame object masks $\{M^{(o)}\}_{o\in\mathcal{O}}$ from \cref{sec:vel_cond}, the estimator $\Phi_{\mathrm{track}}$ propagates all masks jointly through the video using SAM2~\cite{ravi2024sam}, which natively handles multi-object propagation and overlap resolution.
Thanks to SAM2's internal memory bank, all historical frames are processed incrementally with constant per-step cost.
Each tracked object is then painted with a distinct color drawn without replacement from a fixed $K$-color palette of maximally separated RGB values (we use $K{=}10$), on a black background, yielding $c^{\mathrm{track}\vphantom{\mathrm{pk}}}_t \in [0,1]^{H\times W\times 3}$.

\topic{Online Memory Update During Sampling}
\label{sec:online_update}
During autoregressive sampling, the model generates one latent frame at a time, where each latent frame decodes to four pixel frames under the Wan VAE's temporal upsampling.
After each new latent frame $\ell$ is committed, we decode it to pixel space, run both estimators on the new frames, and encode the resulting condition maps back to latent space:
{
\begin{equation}
c^{\mathrm{pos}\vphantom{\mathrm{pk}}}_{\ell} = \Phi_{\mathrm{pos}}(\hat{x}_{\leq\ell}),\;\;
c^{\mathrm{track}\vphantom{\mathrm{pk}}}_{\ell} = \Phi_{\mathrm{track}}(\hat{x}_{\leq\ell},\,\{M^{(o)}\}),
\label{eq:online_update}
\end{equation}
}
where $\hat{x}_{\leq\ell}$ denotes all decoded pixel frames up to and including latent frame $\ell$.
Although both estimators conceptually receive the full history, each component operates incrementally: the Wan VAE's causal temporal convolutions decode and encode only the new latent frame using cached features from previous frames; $\Phi_{\mathrm{pos}}$ estimates depth from only the most recent $L$ frames (\cref{sec:pos_map}); and $\Phi_{\mathrm{track}}$ leverages SAM2's memory bank.
The per-step cost of the entire online memory update is therefore constant regardless of the total video length.

\topic{Teacher-Forcing Training}
\label{sec:teacher_forcing}
Stage~2 is trained in a Teacher-Forcing manner with causal attention.
At each training step, the model receives a ground-truth video $x_{0:N}$ and the corresponding ground-truth conditions $c^{\Delta v\vphantom{\mathrm{pk}}}_{0:N}$, $c^{\mathrm{pos}\vphantom{\mathrm{pk}}}_{0:N}$, $c^{\mathrm{track}\vphantom{\mathrm{pk}}}_{0:N}$.
Each frame $x_i$ is denoised while attending only to the clean ground-truth context of all preceding frames:
{
\begin{equation}
\hat{v}_i \;=\; v_\theta\!\left(z^{\mathrm{noisy}}_i, \tau, \; x_0, x_{1:i-1}^{\mathrm{gt}}, \; c^{\Delta v\vphantom{\mathrm{pk}}}_{<i}, \, c^{\mathrm{pos}\vphantom{\mathrm{pk}}}_{<i}, \, c^{\mathrm{track}\vphantom{\mathrm{pk}}}_{<i}\right),
\label{eq:tf_forward}
\end{equation}
}
where $x_{1:i-1}^{\mathrm{gt}}$ denotes clean ground-truth latents provided as context (not the model's own predictions) and $\tau$ is the diffusion timestep.
The causal attention mask ensures that frame $i$ cannot attend to any frame $j > i$, while the temporal shift of the condition channels (\cref{sec:condition_injection}) ensures that each condition slot carries information strictly from the previous frame.

We adopt Teacher-Forcing~\cite{williams1989learning,jin2024pyramidal} with supervised finetuning rather than distillation~\cite{yin2025slow,huang2025self,zhu2026causal} mainly for a practical reason: Stage~2 must learn two new condition branches ($c^{\mathrm{pos}\vphantom{\mathrm{pk}}}$, $c^{\mathrm{track}\vphantom{\mathrm{pk}}}$) that no bidirectional teacher has seen, and rollout-based objectives (e.g., Self-Forcing~\cite{huang2025self}) would have to run the online estimators inside every training rollout. Teacher-Forcing is not irreplaceable, however: we compare it against Diffusion-Forcing and Self-Forcing trained under the same budget and find it best overall (see \cref{sec:supp_paradigm}).

\subsection{Condition Injection via Shifted Channel Concatenation}
\label{sec:condition_injection}

\topic{Latent Preparation}
\label{sec:latent_prep}
We encode each condition map with the pretrained Wan VAE $\mathcal{E}$.
The resulting condition latents $z^{\Delta v\vphantom{\mathrm{pk}}}$, $z^{\mathrm{pos}\vphantom{\mathrm{pk}}}$, and $z^{\mathrm{track}\vphantom{\mathrm{pk}}}$ all share the spatio-temporal shape of the noisy video latent $z^{\mathrm{noisy}}$.

\topic{Shifted Channel Concatenation}
\label{sec:concat}
The Wan2.2-TI2V-5B variant conditions on the first frame by fusing its clean VAE latent directly into the first temporal slot of the noisy latent: during the denoising process, the first latent frame is always held at the clean encoded value of $x_0$, ensuring that the generated video is anchored to the input image.
The augmented DiT input concatenates all condition latents along the channel dimension after a one-frame forward shift (with the first slot zeroed):
{
\begin{equation}
\tilde{z} = \mathrm{Concat}\!\left(z^{\mathrm{noisy}},\, \mathrm{Shift}(z^{\Delta v\vphantom{\mathrm{pk}}}),\, \mathrm{Shift}(z^{\mathrm{pos}\vphantom{\mathrm{pk}}}),\, \mathrm{Shift}(z^{\mathrm{track}\vphantom{\mathrm{pk}}})\right),
\label{eq:concat}
\end{equation}
}
where $\mathrm{Shift}(\cdot)$ denotes the one-frame forward shift along the latent time axis.
The temporal shift ensures that the condition aligned with latent frame $\ell$ is always derived from the previous latent frame's content, so under causal attention, no in-frame information leaks from $x_{\ell}$ into the conditioning at $\ell$.
The DiT's patch-embedding layer is split into a pretrained branch on the original $z^{\mathrm{noisy}}$ channels (initialized from the backbone weights) and zero-initialized branches on each new condition stream; their token-space outputs are summed before the stacked DiT blocks.
Zero-initialization guarantees that the augmented model is numerically identical to the pretrained backbone at the start of finetuning, after which the conditional branches gradually grow to incorporate the new signals.

%% file: sections/04_experiments.tex
\input{tables/main_results}
\begin{figure}[t]
    \centering
    \includegraphics[width=0.9\linewidth]{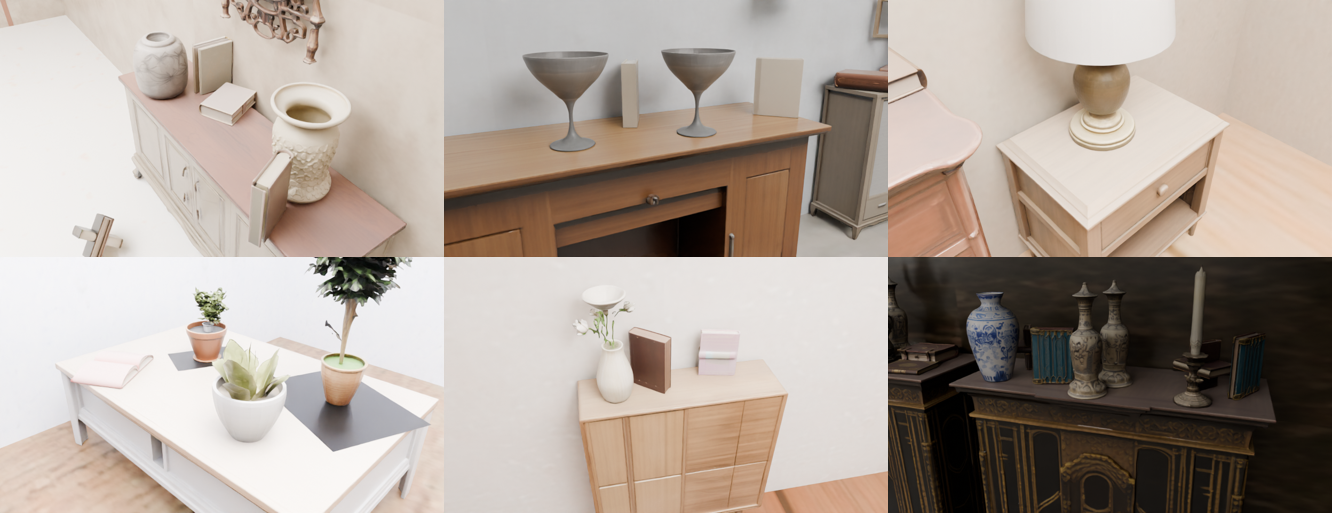}
    \caption{Representative scenes from our curated rigid-body dataset.} 
    \label{fig:dataset}
\end{figure}
\input{tables/wild_results}
\section{Experiments}
\label{sec:experiments}

\subsection{Implementation Details}
\label{sec:impl_details}

\topic{Datasets}
\label{sec:datasets}
We curate our training and evaluation data on SAGE~\cite{xia2026sage}, a large-scale corpus of $10$k pre-generated indoor scenes.
We focus on tabletop rigid-body dynamics involving collisions, frictional contact, and tumbling of small objects.
For each scene, dynamic objects are filtered to keep the resulting dynamics within a tractable complexity range, and the user-specified events $\mathcal{U}$ in \cref{eq:user_events} are randomly sampled by a fixed set of rules.
Multi-body dynamics are simulated with a lightweight PyBullet~\cite{coumans2016pybullet} pipeline, and the frames are rendered with Blender~\cite{blender2018}.
Each video has $49$ frames at $832\!\times\!480$ resolution.
In total, we render approximately $100$k videos, with $3$k held out for validation and evaluation (primarily for constructing FVD reference distributions), and the remainder is used for training.
Representative examples are shown in \cref{fig:dataset}; we refer the reader to \cref{sec:supp_dataset} for further dataset construction details.

\subsection{Evaluation on Synthetic Data}
\label{sec:eval_main}
We evaluate PhysStream on the proposed synthetic benchmark for physics-grounded image-to-video generation.

\topic{Baselines and Settings}
\label{sec:baselines}
We select all methods from \cref{tab:desiderata} that support image-to-video generation and whose control condition can be aligned with our velocity-increment signal, yielding seven baselines:
Force Prompting~\cite{gillman2025force},
PhysCtrl~\cite{wang2025physctrl},
DragAnything~\cite{wu2024draganything},
Tora~\cite{zhang2025tora},
FlashMotion~\cite{li2026flashmotion},
DragStream~\cite{zhou2025streaming},
and RealWonder~\cite{liu2026realwonder}.
We organize the evaluation into two test sets:
(i)~$64$ videos with a single velocity increment on one object at frame~$0$, for baselines that do not support scene-level or mid-frame control (DragAnything, Force Prompting and PhysCtrl);
(ii)~$64$ videos sampled from the standard dataset with multi-object interactive control, for all remaining baselines.

\input{tables/ablation}
\input{tables/posmap_depth}

\topic{Metrics}
\label{sec:metrics}
We evaluate generation quality with eight metrics organized into three groups.

\textit{General physical correctness.}
We use \textbf{FVD}~\cite{unterthiner2018towards,skorokhodov2022stylegan} and \textbf{FVMD}~\cite{liu2024fr} to measure how well the distribution of generated videos matches the simulated ground truth. FVD embeds each video with an I3D network pretrained on Kinetics-400 and computes the Fr\'echet distance between the feature distributions of generated and ground-truth videos, capturing overall distributional similarity; FVMD replaces appearance features with motion features---velocity and acceleration histograms of tracked points---and therefore focuses specifically on motion-pattern similarity.

\textit{Fine-grained motion accuracy.}
We use CoTracker3~\cite{karaev2025cotracker3} to track $32$ query points sampled on each dynamic object in both the ground-truth and generated videos, and report three trajectory-level metrics: \textbf{traj-ADE} (average pixel-distance error between predicted and ground-truth tracks), \textbf{traj-ADE-median} (a more robust median variant), and \textbf{failure rate} (fraction of tracked points in the generated video that either lose track or deviate by more than $30$\,px from the ground truth---a deliberately strict threshold).

\textit{Consistency.}
We report three complementary consistency metrics.
\textbf{Scene consistency} is the subject consistency metric from VBench++~\cite{huang2025vbench++}, capturing overall temporal coherence of the generated scene.
\textbf{Object consistency} is our modified metric that uses SAM2 to track and crop each dynamic object individually, computing per-object appearance consistency---this is motivated by our static-camera setting where per-object motion quality is more informative than whole-frame metrics.
\textbf{Photometric consistency} follows WorldScore~\cite{duan2025worldscore} and measures forward--backward optical-flow agreement.

More details on metric choices and modifications are provided in \cref{sec:supp_metrics}.

\topic{Results}
\label{sec:main_results}
Quantitative results are shown in \cref{tab:main_results}.
On test set~(ii), \name outperforms all baselines on the physics-sensitive metrics: FVMD, traj-ADE, traj-ADE-median, and failure rate consistently show that our generated dynamics more closely follow the ground-truth physical motion, and consistency scores are near-optimal across the board.

\begin{figure*}[tp]
    \centering
    \includegraphics[width=\textwidth]{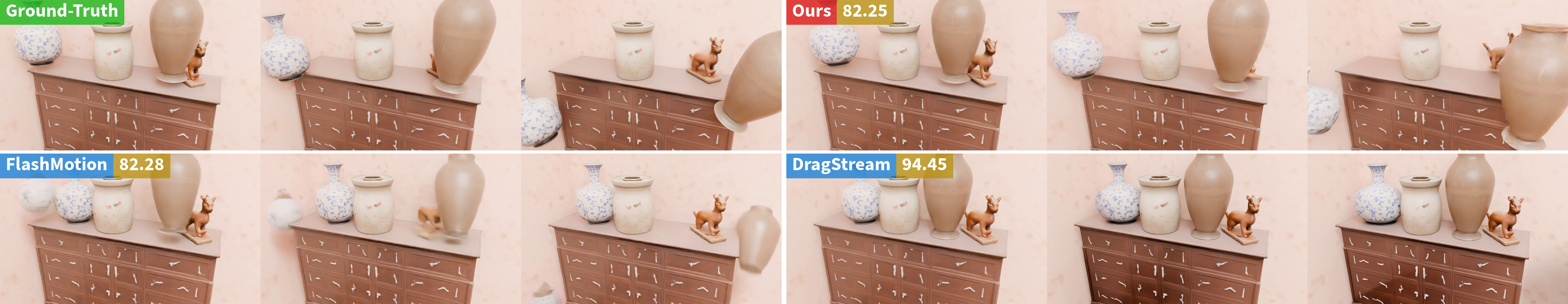}
    \caption{Limitation of consistency metrics. Numbers show the average of scene, object, and photometric consistency. FlashMotion scores comparably to ours but produces visible artifacts and hallucinated objects; DragStream generates a nearly static scene yet achieves the highest consistency score.}
    \label{fig:consistency_limitation}
\end{figure*}

\topic{Limitation of Consistency Metrics}
\label{sec:consistency_limitation}
While consistency metrics are important for evaluating video generation quality, we note that they can be inflated by degenerate generations where objects remain nearly static or drift rigidly in pixel space without physically plausible dynamics: such outputs trivially preserve appearance consistency, leading to artificially high scores.
This phenomenon is illustrated in \cref{fig:consistency_limitation}.

\subsection{Evaluation on In-the-Wild Data}
\label{sec:eval_wild}
To assess generalization beyond the synthetic training distribution, we evaluate \name in three settings: (i)\textbf{~In-the-wild scenes:} 20 input images paired with velocity-increment signals randomly generated under a fixed set of rules, compared against the four baselines that support full interactive control; (ii)\textbf{~Real-world captures}: $16$ cluttered indoor scenes from OCID~\cite{suchi2019ocid} and $10$ real videos with ground truth from the Physics-IQ benchmark~\cite{motamed2025physicsiq}; and (iii)~\textbf{Non-rigid objects}: Two kinds of scenes where the same control and scene-memory paradigm is applied to deformable balls and cloth.

\topic{Metrics}
\label{sec:wild_setting}
Since no ground-truth video is available for in-the-wild inputs (except for the Physics-IQ benchmark), most metrics from \cref{sec:metrics} cannot be applied.
We therefore adopt an MLLM evaluation for all settings: following VideoPhy~\cite{bansal2024videophy, wang2025physctrl}, we query GPT-4o for \textbf{Semantic Adherence (SA)} and \textbf{Physical Commonsense (PC)} scores on a 1--5 Likert scale.
For setting~(i), we additionally report \textit{human preference}: evaluators are shown the five results (four baselines and ours) side by side and asked to select the best one along three axes: \textbf{physical plausibility} (Phys.), \textbf{motion accuracy} (Motn.), and \textbf{visual quality} (Vis.), reported as win rate~(\%).
More details are provided in \cref{sec:supp_wild}.

\input{tables/generalization_pair}

\begin{figure*}[tp]
    \centering
    \includegraphics[width=\textwidth]{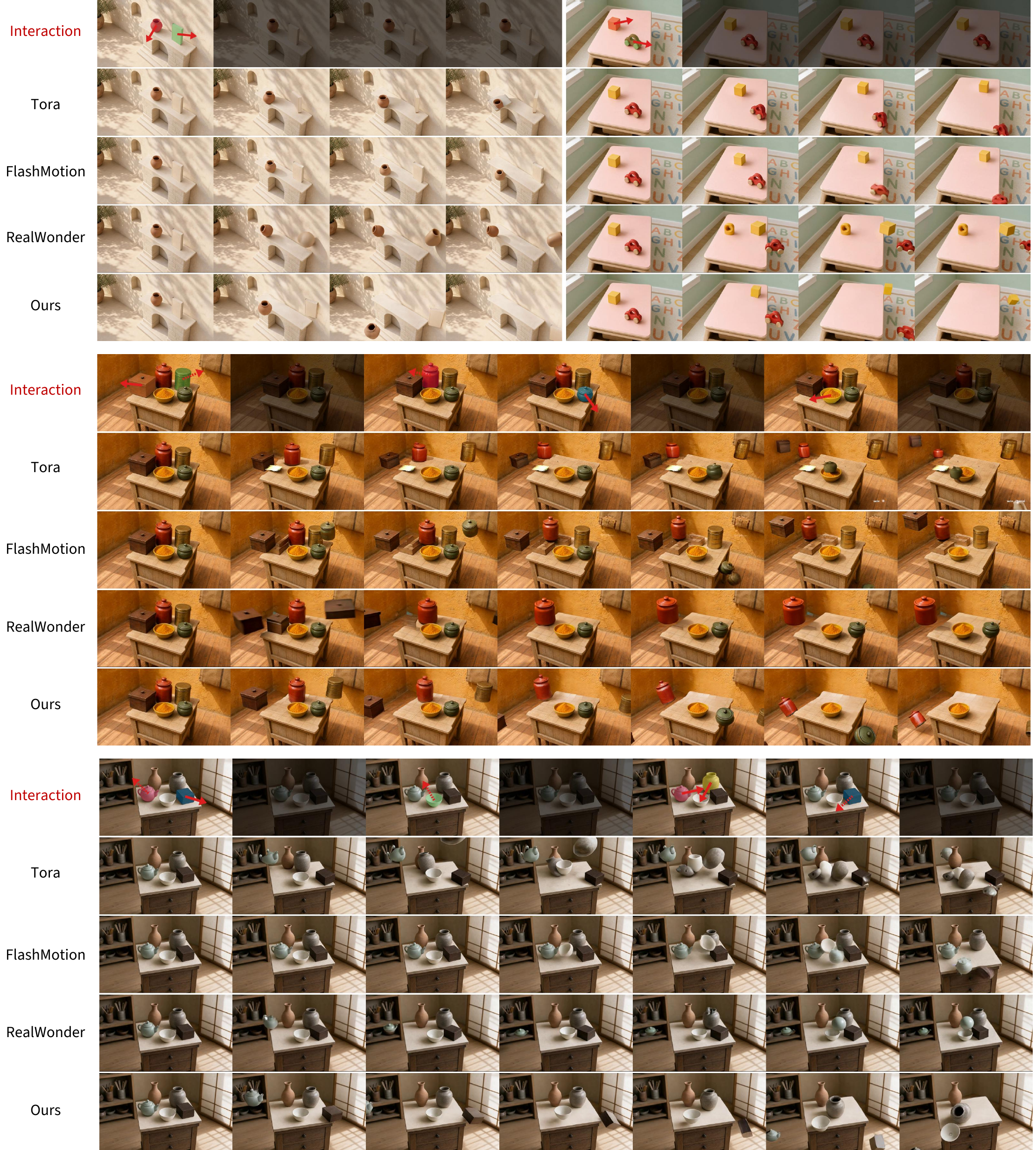}
    \caption{Qualitative comparison on multi-object rigid-body scenes. Compared with baselines, our method achieves physics-grounded video generation with multi-object interactions, while baselines produce distorted geometries and inconsistent motions.}
    \label{fig:qualitative}
\end{figure*}

\begin{figure*}[tp]
    \centering
    \includegraphics[width=\textwidth]{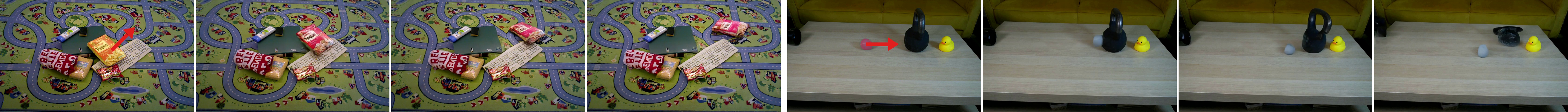}
    \caption{Results on real-world captures. \textbf{Left:} a cluttered indoor scene from OCID; the food package is pushed and correctly thrown across the clutter. \textbf{Right:} a Physics-IQ scenario where a rolling ball hits a weight placed in front of a duck; despite appearance drift on this out-of-distribution input, the ball--weight collision is modeled correctly and the duck is protected as in the real video. The first panel of each case shows the input with the applied velocity increment. Input frames \copyright~TU Wien ACIN (OCID) and Google DeepMind \& INSAIT (Physics-IQ), CC~BY~4.0.}
    \label{fig:realworld}
\end{figure*}

\begin{figure*}[tp]
    \centering
    \includegraphics[width=\textwidth]{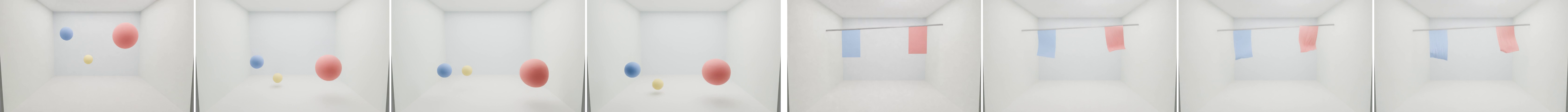}
    \caption{Results of non-rigid dynamics generated by the finetuned models: \textbf{Left:} Elastically bouncing balls. \textbf{Right:} Fluttering cloth.}
    \label{fig:nonrigid}
\end{figure*}

\topic{Results}
\label{sec:wild_results}
(i)~\Cref{tab:wild_results} reports quantitative results and \cref{fig:qualitative} shows representative examples.
\name achieves a clear advantage across all five metrics: both MLLM scores are the highest, and human evaluators prefer our results in over $80$\% of comparisons on every axis---indicating that the quality gap over baselines is substantial and consistent in general in-the-wild scenarios.
(ii)~\Cref{tab:realworld} and \cref{fig:realworld} show the results on real-world captures. SA and PC remain as high as in setting~(i) on the heavily cluttered OCID scenes, and on Physics-IQ \name additionally reaches an official score of $47.9$ on the selected solid-mechanics subset, where the initial velocity of the moving object is derived from the real clip; the generated motion follows the real direction and collision timing, with the object speed as the main remaining discrepancy.
(iii)~\Cref{tab:nonrigid} and \cref{fig:nonrigid} show that non-rigid materials transfer well under the same condition paradigm: the same velocity-increment control and structured scene memory, without any change to the method, drive deformable balls to bounce elastically and cloth to fold and flutter, with SA/PC on par with the rigid-body results. These results are obtained by finetuning our full model on a small synthetic dataset built for each material ($10$k clips each; $5$k iterations), suggesting that extending \name to richer materials mainly requires extending the dataset.

\input{tables/longhorizon}

\begin{figure*}[tp]
    \centering
    \includegraphics[width=\textwidth]{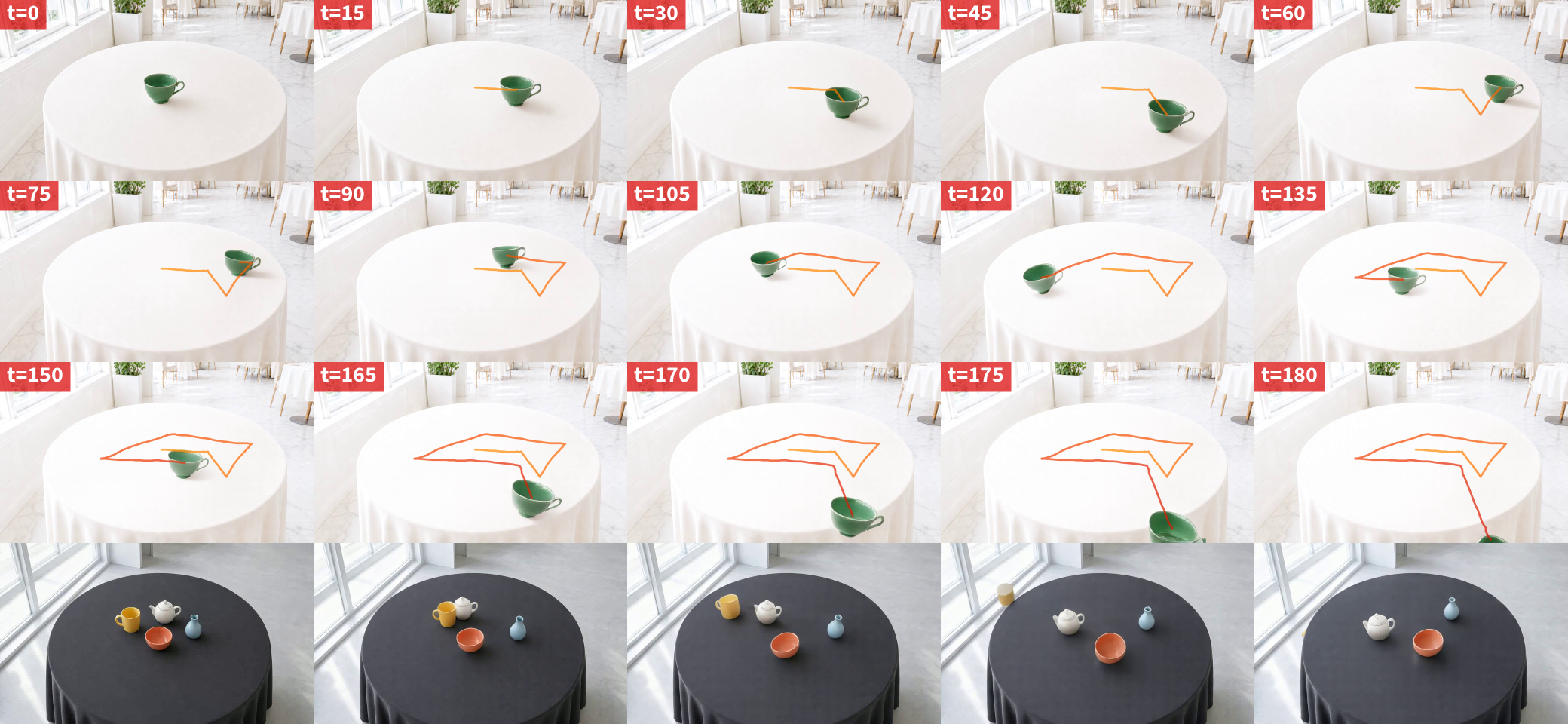} 
    \caption{Long-horizon generation well beyond the $49$-frame training window. \textbf{Top:} a jade-colored teacup performs a random walk on a tabletop, consistently following the randomly injected velocity-increment interactions and preserving its appearance until it falls off the table edge at frame~$180$. \textbf{Bottom:} a more complex multi-object case from our long-horizon benchmark, where every object receives periodic velocity increments.}
    \label{fig:longvideo}
\end{figure*}

\subsection{Long Video Generation}
\label{sec:longvideo}
Although \name is trained on $49$-frame clips for both stages, the autoregressive structure and the strict use of historical scene memory together permit straightforward extension to longer horizons without any architectural change.
To quantify this, we build a long-horizon control benchmark of $5$ multi-object tabletop scenes with $301$ frames ($6\times$ the training horizon) and interactions throughout (see \cref{sec:supp_lh_metrics} for details), and report per-segment results in \cref{tab:longhorizon}. Consistency decreases gradually over the horizon due to accumulated appearance drift---the well-known failure mode of autoregressive generation---yet the response rate and control accuracy remain high throughout: drift degrades appearance, not the model's ability to respond to control signals. \Cref{fig:longvideo} shows two representative sequences.

\subsection{Ablation Study}
\label{sec:ablation}

\topic{Structured Scene Memory}
We conduct ablation experiments on the $64$ test cases from test set~(ii) in \cref{sec:baselines}.
To reduce variance across training checkpoints, we average results over the last $10$ saved checkpoints.
We evaluate five configurations:
\textit{(a) Stage~1 only} (velocity-increment condition only, bidirectional);
\textit{(b) Stage~2 with velocity only} (the same condition, but in causal autoregressive form);
\textit{(c) Stage~2 with velocity + positional map};
\textit{(d) Stage~2 with velocity + tracking map};
\textit{(e) \name full} (Stage~2 with velocity, positional map, and tracking map).
Configuration (a) does not support on-the-fly interactive control: its motion control must be specified in advance.
\Cref{tab:ablation} reports the results.
Our full model~(e) achieves the best or near-best scores on nearly all metrics.
The one exception is FVD, where fewer conditions yield slightly better scores; this is expected because FVD measures distributional similarity to the training set, and in our i.i.d.\ setting, the unconditional model fits this distribution most directly---additional conditions require longer convergence, so a small gap under equal training time is reasonable.
Beyond the per-metric comparison, all autoregressive configurations~(b--e) substantially outperform the bidirectional Stage-1 model~(a), whose training has already converged, confirming that causal models are better suited to our task where physical dynamics are inherently causal.
The benefit of structured scene memory extends well beyond the numeric margins in \cref{tab:ablation}: our randomly sampled test set does not cover many challenging corner cases.
To isolate the effect of the positional map, we identify the per-object subset most sensitive to 3D geometry---objects whose ground-truth depth displacement is largest---and compute the best per-object Traj-ADE across all checkpoints for configurations~(b) and~(c).
As \cref{tab:posmap_depth} shows, the positional map provides a steadily increasing advantage as depth motion grows, reaching \textbf{15.5\%} for the top-10\% objects; on the full set the margin is modest ($2.1$\%), confirming that the positional map primarily aids geometrically challenging motions.

\begin{figure*}[tp]
    \centering
    \includegraphics[width=\textwidth]{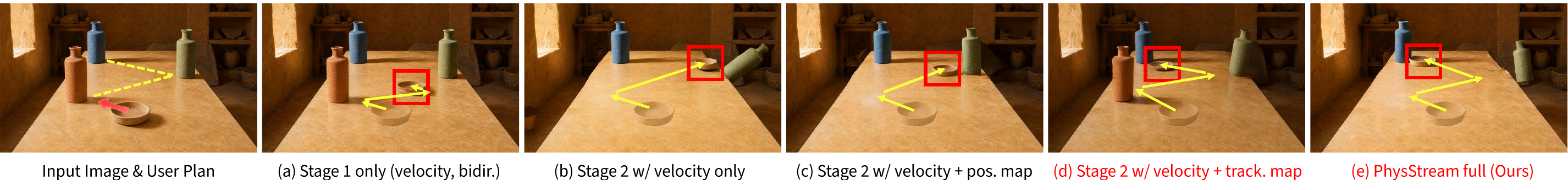}
    \caption{Ablation on the zig-zag test case from \cref{fig:teaser} (top row). All five configurations receive the same user interaction (shown in the leftmost panel). Only configurations with the tracking map---(d) and (e)---complete the full zig-zag trajectory. (a)~produces imprecise, drifting control; (b) and (c) get stuck at the final turn, unable to redirect the object once it has moved far from its first-frame mask.}
    \label{fig:ablation}
\end{figure*}

\Cref{fig:ablation} further illustrates the effect of the tracking map on the zig-zag test case from the teaser (\cref{fig:teaser}, top row), where a ceramic dish must execute multiple sharp turns to strike successive vases.
Using the same user interaction across all ablation configurations, we find that only models equipped with the tracking map---configurations~(d) and~(e)---successfully complete the full zig-zag trajectory.
Configuration~(a) produces imprecise control with the object drifting off course, while configurations~(b) and~(c) get stuck at the final turn, unable to redirect the object once it has moved far from its first-frame mask position.
We further quantify the importance of the tracking map on the long-horizon benchmark of \cref{sec:longvideo} by dropping the estimated tracking map at inference; see \cref{sec:supp_track_long} for the quantitative results and analysis.

\input{tables/runtime}
\input{tables/runtime_speed}

\subsection{Runtime Analysis}
\label{sec:runtime}
Runtime is a crucial practical consideration for interactive video generation, especially since we add two online estimators to the generation loop. \Cref{tab:runtime} breaks down the unaccelerated system: the $50$-step denoising dominates ($68\%$), followed by Depth-Anything-3 ($19\%$), the VAE ($8\%$; incremental decoding plus re-encoding of the two memory conditions), and SAM2 ($4\%$). Both dominant costs are readily reducible: (a)~we distill our model into a $4$-step causal generator with distribution matching distillation~\cite{yin2024one}, halving the end-to-end latency with less than $1\%$ average metric degradation on the synthetic benchmark, and (b)~we replace the depth estimator with a $4\times$ smaller metric-depth model, estimating the intrinsics once on the first frame (the camera is static). Together with I/O-level engineering of the estimation loop, the system runs $3.4\times$ faster at lower memory (\cref{tab:runtime_speed}). The remaining budget is dominated by the VAE round trips, which efficient or VAE-free video generators are designed to remove; combined with the trend towards real-time online estimators, real-time rates appear within reach and are left as future work.

%% file: tables/main_results.tex
\begin{table*}[!tp]
    \caption{Quantitative comparison on synthetic data. (i): single-object with first-frame control; (ii): multi-object with interactive control.
    $^{*}$Tora and FlashMotion do not support interactive control; we strengthen their setting by providing the ground-truth center-of-mass trajectory as control input.
    For RealWonder we skip scene reconstruction and directly use the ground-truth scene.
    Higher is better~($\uparrow$); lower is better~($\downarrow$). Here we include consistency-based metrics from VBench for completeness, we discuss their limitations at the end of \cref{sec:eval_main}.}
    \label{tab:main_results}
    \centering
    \small
    \setlength{\tabcolsep}{4pt}
    \renewcommand{\arraystretch}{1.1}
    \begin{tabular}{cl|cc|ccc|ccc}
        \toprule
         &  & FVD\,$\downarrow$ & FVMD\,$\downarrow$ & Traj-ADE\,$\downarrow$ & Traj-ADE-M\,$\downarrow$ & Failure\,$\downarrow$ & Scene Cons.\,$\uparrow$ & Obj. Cons.\,$\uparrow$ & Photo. Cons.\,$\uparrow$ \\
        \midrule
        \multirow{4}{*}{(i)}
        & DragAnything & 1084 & 41315 & \cellcolor{Goldenrod}97.47 & \cellcolor{Goldenrod}81.78 & \cellcolor{Goldenrod}69.70 & 88.00 & \cellcolor{YellowOrange}94.11 & 35.65 \\
        & Force Prompting & \cellcolor{Goldenrod}606.6 & \cellcolor{Goldenrod}2142 & 105.3 & \cellcolor{LightYellow1}94.59 & 74.13 & \cellcolor{LightYellow1}94.65 & 87.70 & \cellcolor{LightYellow1}80.72 \\
        & PhysCtrl & \cellcolor{LightYellow1}626.8 & \cellcolor{LightYellow1}3344 & \cellcolor{LightYellow1}104.9 & 94.95 & \cellcolor{LightYellow1}73.91 & \cellcolor{YellowOrange}97.93 & \cellcolor{Goldenrod}92.42 & \cellcolor{YellowOrange}93.45 \\
        & \name (Ours) & \cellcolor{YellowOrange}492.9 & \cellcolor{YellowOrange}846.0 & \cellcolor{YellowOrange}49.37 & \cellcolor{YellowOrange}33.78 & \cellcolor{YellowOrange}48.11 & \cellcolor{Goldenrod}96.30 & \cellcolor{LightYellow1}87.80 & \cellcolor{Goldenrod}83.24 \\
        \midrule
        \multirow{5}{*}{(ii)}
        & Tora$^{*}$ & \cellcolor{Goldenrod}428.1 & \cellcolor{LightYellow1}1463 & 66.79 & 57.11 & 72.00 & 91.37 & 78.37 & 72.62 \\
        & FlashMotion$^{*}$ & 526.7 & 3751 & \cellcolor{Goldenrod}45.67 & \cellcolor{Goldenrod}39.11 & \cellcolor{Goldenrod}44.28 & \cellcolor{YellowOrange}96.79 & \cellcolor{Goldenrod}86.34 & \cellcolor{YellowOrange}83.25 \\
        & DragStream & 758.2 & 2662 & 70.27 & 63.56 & \cellcolor{LightYellow1}64.67 & 90.46 & \cellcolor{YellowOrange}91.09 & 30.22 \\
        & RealWonder & \cellcolor{LightYellow1}438.8 & \cellcolor{Goldenrod}1183 & \cellcolor{LightYellow1}60.91 & \cellcolor{LightYellow1}49.84 & 64.78 & \cellcolor{LightYellow1}95.03 & 80.94 & \cellcolor{LightYellow1}73.48 \\
        & \name (Ours) & \cellcolor{YellowOrange}413.7 & \cellcolor{YellowOrange}787.0 & \cellcolor{YellowOrange}40.24 & \cellcolor{YellowOrange}32.00 & \cellcolor{YellowOrange}43.15 & \cellcolor{Goldenrod}96.57 & \cellcolor{LightYellow1}85.29 & \cellcolor{Goldenrod}81.72 \\
        \bottomrule
    \end{tabular}
\end{table*}

%% file: tables/wild_results.tex
\begin{table}[h] 
    \caption{Evaluation on in-the-wild data. SA/PC: Semantic Adherence / Physical Commonsense from VideoPhy~\cite{bansal2024videophy} (1--5 Likert); Phys./Motn./Vis.: human preference win rate (\%).}
    \label{tab:wild_results}
    \centering\small
    \setlength{\tabcolsep}{3pt}
    \begin{tabular}{l|cc|ccc}
        \toprule
        & SA\,$\uparrow$ & PC\,$\uparrow$ & Phys.\,$\uparrow$ & Motn.\,$\uparrow$ & Vis.\,$\uparrow$ \\
        \midrule
        Tora & 4.35 & 3.30 & 1.8\% & 2.6\% & 2.2\% \\
        FlashMotion & 4.85 & 3.65 & 4.8\% & 7.0\% & 4.6\% \\
        DragStream & 4.35 & 2.45 & 0.4\% & 0.2\% & 0.4\% \\
        RealWonder & 4.65 & 3.25 & 1.2\% & 1.8\% & 1.6\% \\
        \name (Ours) & \textbf{5.00} & \textbf{4.15} & \textbf{91.8\%} & \textbf{88.4\%} & \textbf{91.2\%} \\
        \bottomrule
    \end{tabular}
\end{table}

%% file: tables/ablation.tex
\begin{table*}[!tp]
    \caption{Ablation results on test set (ii). See \cref{sec:ablation} for configuration definitions and \cref{tab:main_results} for column abbreviations.}
    \label{tab:ablation}
    \centering \small
    \setlength{\tabcolsep}{4pt}
    \renewcommand{\arraystretch}{1.1}
    \begin{tabular}{l|cc|ccc|ccc}
        \toprule
         & FVD\,$\downarrow$ & FVMD\,$\downarrow$ & Traj-ADE\,$\downarrow$ & Traj-ADE-M\,$\downarrow$ & Failure\,$\downarrow$ & Scene Cons.\,$\uparrow$ & Obj. Cons.\,$\uparrow$ & Photo. Cons.\,$\uparrow$ \\
        \midrule
        (a) Stage~1 only (velocity, bidir.) & 424.7 & 1015 & 48.78 & 41.96 & 53.92 & 94.59 & 82.90 & 65.81 \\
        (b) Stage~2 w/\ velocity only & \cellcolor{YellowOrange}399.2 & 941.3 & \cellcolor{YellowOrange}43.75 & \cellcolor{Goldenrod}35.85 & \cellcolor{YellowOrange}47.70 & 96.00 & \cellcolor{Goldenrod}83.80 & \cellcolor{LightYellow1}78.76 \\
        (c) Stage~2 w/\ velocity + pos.\ map & \cellcolor{LightYellow1}399.7 & \cellcolor{LightYellow1}924.7 & 45.04 & 36.79 & \cellcolor{LightYellow1}48.95 & \cellcolor{Goldenrod}96.04 & \cellcolor{LightYellow1}83.78 & 78.66 \\
        (d) Stage~2 w/\ velocity + track.\ map & \cellcolor{Goldenrod}399.5 & \cellcolor{Goldenrod}910.5 & \cellcolor{LightYellow1}44.54 & \cellcolor{LightYellow1}36.56 & 49.50 & \cellcolor{LightYellow1}96.02 & 83.72 & \cellcolor{Goldenrod}79.56 \\
        (e) \name full (Ours) & 404.8 & \cellcolor{YellowOrange}879.8 & \cellcolor{Goldenrod}43.82 & \cellcolor{YellowOrange}35.58 & \cellcolor{Goldenrod}47.73 & \cellcolor{YellowOrange}96.21 & \cellcolor{YellowOrange}84.20 & \cellcolor{YellowOrange}80.37 \\
        \bottomrule
    \end{tabular} 
\end{table*}

%% file: tables/posmap_depth.tex
\begin{table}[t]
    \caption{Per-object best Traj-ADE grouped by GT depth displacement. Top-$k$\% selects the $n$ objects with the largest depth change.}
    \label{tab:posmap_depth}
    \centering\small
    \setlength{\tabcolsep}{4pt}
    \begin{tabular}{l|cccc}
        \toprule
         & Full$_{\scriptscriptstyle n=177}$ & Top\,50\%$_{\scriptscriptstyle n=88}$ & Top\,20\%$_{\scriptscriptstyle n=35}$ & Top\,10\%$_{\scriptscriptstyle n=17}$ \\
        \midrule
        w/o pos.\ map & 11.5 & 16.5 & 20.6 & 25.4 \\
        w/ pos.\ map  & \textbf{11.2} & \textbf{15.7} & \textbf{18.3} & \textbf{21.5} \\
        \midrule
        Gain & $+2.1$\% & $+4.9$\% & $+11.5$\% & \textbf{+15.5}\% \\
        \bottomrule
    \end{tabular}
\end{table}

%% file: tables/generalization_pair.tex
\begin{table}[t]
    \begin{minipage}[t]{0.585\columnwidth}
        \centering\small 
        \caption{Evaluation on real-world captures. \textbf{P-IQ:} the official Physics-IQ score on the solid-mechanics subset.}
        \label{tab:realworld}
        \setlength{\tabcolsep}{3.5pt}
        \begin{tabular}{l|cc|c}
            \toprule
& SA\,$\uparrow$ & PC\,$\uparrow$ & P-IQ\,$\uparrow$ \\
            \midrule
OCID & 5.00 & 4.06 & N/A \\
Physics-IQ & 4.80 & 3.70 & 47.86 \\
            \bottomrule
        \end{tabular}
    \end{minipage}\hfill
    \begin{minipage}[t]{0.375\columnwidth}
        \centering\small 
        \caption{Evaluation on non-rigid dynamics tested for the finetuned models.}
        \label{tab:nonrigid}
        \setlength{\tabcolsep}{3.5pt}
        \begin{tabular}{l|cc}
            \toprule
& SA\,$\uparrow$ & PC\,$\uparrow$ \\
            \midrule
Balls & 4.20 & 4.20 \\
Cloth & 5.00 & 5.00 \\
            \bottomrule
        \end{tabular}
    \end{minipage}
\end{table}

%% file: tables/longhorizon.tex
\begin{table}[t]
    \caption{Long-horizon control benchmark, evaluated per $100$-frame segment: average consistency, the fraction of control events the target object responds to, and the directional agreement of the response with the command. Metric definitions are in \cref{sec:supp_lh_metrics}.}
    \label{tab:longhorizon}
    \centering\small
    \setlength{\tabcolsep}{2.5pt}
    \begin{tabular}{l|ccc}
        \toprule
Frames & Avg.\ Consistency\,$\uparrow$ & Successful Respond\,$\uparrow$ & Control Accuracy\,$\uparrow$ \\
        \midrule
1--100   & 95.53 & 100.0\% & 95.2\% \\
101--200 & 92.29 & 92.5\%  & 87.1\% \\
201--300 & 87.18 & 87.7\%  & 73.7\% \\
        \bottomrule
    \end{tabular}
\end{table}

%% file: tables/runtime.tex
\begin{table}[t]
    \caption{Per-component latency of the unaccelerated system for one $49$-frame clip at $832\times480$ on a single B6000 GPU.}
    \label{tab:runtime}
    \centering\small
    \setlength{\tabcolsep}{4pt}
    \begin{tabular}{l|cccc}
        \toprule
& 50-step Denoising & VAE & DA3 & SAM2 \\
        \midrule
Latency    & 45.2\,s & 5.5\,s & 12.7\,s & 2.9\,s \\
Percentage & \textbf{68.1\%} & 8.3\% & \textbf{19.2\%} & 4.4\% \\
        \bottomrule
    \end{tabular} 
\end{table}

%% file: tables/runtime_speed.tex
\begin{table}[t]
    \caption{System-level latency, throughput, and peak memory per $49$-frame clip on a single B6000 GPU.}
    \label{tab:runtime_speed}
    \centering\small
    \setlength{\tabcolsep}{5pt}
    \begin{tabular}{l|ccc}
        \toprule
& Latency\,$\downarrow$ & FPS\,$\uparrow$ & Memory (GB)\,$\downarrow$ \\
        \midrule
Ours                  & 66.3\,s & 0.74 & 56.8 \\
+ DMD                 & 32.9\,s & 1.49 & 56.3 \\
+ DMD \& faster depth & \textbf{19.3\,s} & \textbf{2.54} & \textbf{52.8} \\
        \bottomrule
    \end{tabular} 
\end{table}

%% file: sections/05_conclusion.tex
\section{Conclusion and Limitations}
\label{sec:conclusion}

We presented \name, an autoregressive image-to-video model for physics-grounded interactive generation in tabletop rigid-body scenes.
\name conditions on sparse, user-specified velocity-increment signals that encode physical quantities, together with a structured scene memory---positional maps and object-tracking maps derived online from previously generated frames.
Across synthetic, real-world, and long-horizon settings, \name consistently improves physics-related consistency and motion-control adherence over recent controllable baselines.

That said, \name still struggles with extremely complex motion, particularly tumbling, and its validated scope is limited to rigid-body dynamics, with richer materials currently relying on additional finetuning data. We also leave real-time generation to future work.

%% file: appendix.tex
\section{Two-Stage Training Procedure}
\label{sec:supp_training}

\name is trained in two stages, both using the standard $v$-prediction flow-matching loss~\cite{lipman2022flow}.

\topic{Stage~1: Bidirectional Model with Motion Control.}
We perform full finetuning on the Wan2.2-TI2V-5B~\cite{wan2025wan} backbone with a differential learning-rate schedule.
The new velocity-increment patch embedding layer is trained at $1\!\times\!10^{-4}$; the pretrained attention blocks and time-projection layers at $5\!\times\!10^{-5}$; and the pretrained patch embedding and output head at $2\!\times\!10^{-6}$.
Cross-attention key, value, and normalization layers are kept frozen.
The bidirectional model processes all $N$ frames jointly with full self-attention, conditioned only on the velocity-increment map $c^{\Delta v\vphantom{\mathrm{pk}}}$.

\topic{Stage~2: Causal Model with Structured Scene Memory.}
Starting from the Stage-1 checkpoint, we convert self-attention to causal block-wise attention (one latent frame per block) following Causal-Forcing~\cite{zhu2026causal} and train in a Teacher-Forcing~\cite{williams1989learning, jin2024pyramidal} manner: at each training step, frame $i$ is denoised while attending only to the clean ground-truth latents of frames $0, \dots, i{-}1$.
The pretrained patch embedding, the Stage-1 velocity-increment branch, and the output head are frozen.
The two new structured-scene-memory patch embedding layers ($c^{\mathrm{pos}\vphantom{\mathrm{pk}}}, c^{\mathrm{track}\vphantom{\mathrm{pk}}}$) are trained at $1\!\times\!10^{-4}$, and the remaining DiT layers at $1\!\times\!10^{-5}$.
Both stages are trained on $8{\times}$~H100 GPUs for approximately 30 hours each, using AdamW with default parameters.

\section{Dataset Construction Details}
\label{sec:supp_dataset}

\subsection{Main Rigid-Body Dataset}

Our dataset is built on top of SAGE~\cite{xia2026sage}, a corpus of $10$k pre-generated indoor scenes from 3D-Front.
For each scene, we run a three-stage pipeline: \textbf{process} (scene loading, object filtering, camera placement), \textbf{simulate} (PyBullet~\cite{coumans2016pybullet} multi-body physics), and \textbf{render} (Blender~\cite{blender2018} Cycles with 8 spp + OIDN denoising).

\topic{Object Filtering.}
Dynamic objects resting on each tabletop or cabinet surface are sorted by bounding-box volume; up to $10$ largest objects are kept per surface.
Objects with a minimum bounding-box dimension below $0.15$\,m are excluded as noise.

\topic{Multi-Frame Kick System.}
Rather than a single initial impulse, we apply velocity perturbations at $12$ evenly spaced frame nodes ($t \in \{0, 4, 8, \dots, 44\}$) across the $49$-frame video.
At each node, $0$, $1$, or $2$ kicks are sampled: at frame $0$ the probability is $[50\%, 50\%, 0\%]$ for $[1\text{-kick}, 2\text{-kicks}, \text{skip}]$; at subsequent frames it is $[20\%, 10\%, 70\%]$, keeping most frames purely physics-driven.
Two kick types are used: \textit{kick-A} (horizontal only, $v_{xy} \in [0.5, 1.0]$\,m/s) and \textit{kick-B} (horizontal + upward vertical, $v_{xy} \in [1.0, 1.5]$, $v_z \in [1.0, 1.5]$\,m/s), with a $60\%/40\%$ selection probability when the object is on the floor surface.
Each object may receive at most $3$ kicks with a minimum interval of $8$ frames between consecutive kicks.

\topic{Candidate Selection.}
Before applying a kick, we verify that the target object is (1)~within the camera frustum (at least one AABB corner projects inside the FOV), and (2)~at least $80\%$ visible (via $1000$-ray occlusion check in PyBullet).

\topic{Velocity Semantics.}
Kicks are additive velocity changes ($\mathbf{v}_{\mathrm{new}} = \mathbf{v}_{\mathrm{current}} + \Delta\mathbf{v}$), so a second kick on an already-moving object compounds with existing momentum, producing complex trajectories including tumbling and multi-object collisions.

\topic{Rendering.}
Each frame is rendered at $832\!\times\!480$ with Blender Cycles (CPU, 8 samples, OIDN denoising).
Six aligned modalities are produced per video: RGB, per-object instance mask, velocity-increment canvas (painted on frame-0 mask), normalized positional map (camera-frame coordinates), object-tracking map (palette-colored per-object masks), and inverse depth.
All frames are encoded as lossless FFV1 MKV at 16 fps.

\topic{Camera Placement.}
For each qualifying floor object, $10$ camera groups are sampled with depression angles in $[30^{\circ}, 60^{\circ}]$ and distances in $[0.8, 1.2] \times d_{\min}$, where $d_{\min}$ is the minimum distance to fit the object group within a $90^{\circ}$ FOV.
Cameras are reject-sampled to lie within room bounds (wall margin $1.0$\,m).

\subsection{Deformable-Ball and Cloth Datasets}
\label{sec:supp_deform}
For the non-rigid experiments (\cref{sec:eval_wild}), we build two additional $10$k-clip synthetic datasets with the same resolution, condition rendering, and kick sampling as the main dataset, replacing only the scenes and the simulator. \emph{Deformable balls}: $2$--$3$ elastic balls launched with random initial velocities in a plain box room, simulated with the material point method; \emph{Cloth}: $2$--$3$ cloth pieces hanging from a rod under a gusting wind, simulated with a mass--spring model. Each set holds out $100$ clips for validation. The full model is finetuned on each set for ${\sim}5$k iterations (five epochs) from the final rigid-body checkpoint, with all condition patch-embedding branches frozen.

\section{First-Frame Mask: Experimental Evidence}
\label{sec:supp_mask}

In \cref{sec:mask_choice} we state that using the per-frame (current-position) mask instead of the first-frame mask for the velocity-increment condition degrades generation quality due to information leakage during bidirectional training.

To quantify this effect, we train a lightweight rank-512 LoRA variant for each mask strategy (first-frame mask vs.\ per-frame mask) across both stages, and evaluate on a held-out set of $100$ test videos using five metrics: Traj-ADE ($\downarrow$), Traj-ADE-Median ($\downarrow$), Failure Rate ($\downarrow$), FVD ($\downarrow$), and FVMD ($\downarrow$).

\begin{table}[h]
    \centering\small
    \caption{First-frame mask vs.\ per-frame mask across training stages.}
    \label{tab:mask_ablation}
    \setlength{\tabcolsep}{3pt}
    \begin{tabular}{ll|ccccc}
        \toprule
         & Mask & ADE$\downarrow$ & ADE-M$\downarrow$ & Fail\%$\downarrow$ & FVD$\downarrow$ & FVMD$\downarrow$ \\
        \midrule
        \multirow{2}{*}{Stage 1}
        & first-frame  & 46.3 & 39.0 & 51.8 & 228.1 & 821 \\
        & per-frame    & \textbf{30.8} & \textbf{24.8} & \textbf{37.4} & \textbf{183.1} & \textbf{439} \\
        \midrule
        \multirow{2}{*}{Stage 2}
        & first-frame  & \textbf{43.1} & \textbf{37.0} & \textbf{48.0} & \textbf{194.5} & 582 \\
        & per-frame    & 47.8 & 38.6 & 54.4 & 206.9 & \textbf{563} \\
        \bottomrule
    \end{tabular}
\end{table}

\Cref{tab:mask_ablation} confirms the information-leakage mechanism described in \cref{sec:mask_choice}.
In Stage~1 (bidirectional), the per-frame mask is strictly superior across every metric, achieving a $33\%$ lower ADE and nearly halved FVMD.
This is expected: under bidirectional attention, the per-frame mask reveals the kicked object's current spatial position at each frame where a velocity increment is applied, leaking partial trajectory information into the condition channel.
However, when transitioning to causal autoregressive generation in Stage~2, this advantage reverses sharply.
The per-frame-mask model degrades on four of five metrics, because the causal model can no longer access future-frame masks, so the condition distribution shifts abruptly between Stage~1 and Stage~2, widening the gap between the two training stages.
In contrast, the first-frame-mask model improves consistently from Stage~1 to Stage~2, as its condition distribution remains unchanged across both training regimes.
This validates our design choice of anchoring all velocity-increment events to the first-frame mask.

\section{Positional Map: Window Size and Normalization Anchor}
\label{sec:supp_posmap}

In \cref{sec:pos_map} we estimate the normalized positional map using only the most recent $L{=}4$ pixel frames (one latent frame) and normalize with a scale factor $\rho$ anchored to the first frame.
A natural concern is that subsequent frames may contain position values outside the first frame's range, causing clipping.

We evaluate this on our $64$-video validation set by comparing the per-axis min/max of frame~$0$ against the full-sequence min/max for each video (\cref{tab:posmap_clip}).
Only $0.90$\% of pixels are actually clipped, and the average scale-factor deviation is $0.42$\%, confirming that first-frame normalization introduces negligible distortion under our static-camera setting.

\begin{table}[h]
    \centering\small
    \caption{First-frame normalization analysis on $64$ validation videos.}
    \label{tab:posmap_clip}
    \begin{tabular}{lc}
        \toprule
        Clipped pixels (\%) & 0.90 \\
        Scale-factor deviation (\%) & 0.42 \\
        \bottomrule
    \end{tabular}
\end{table}

\begin{figure*}[t]
    \centering
    \includegraphics[width=\textwidth]{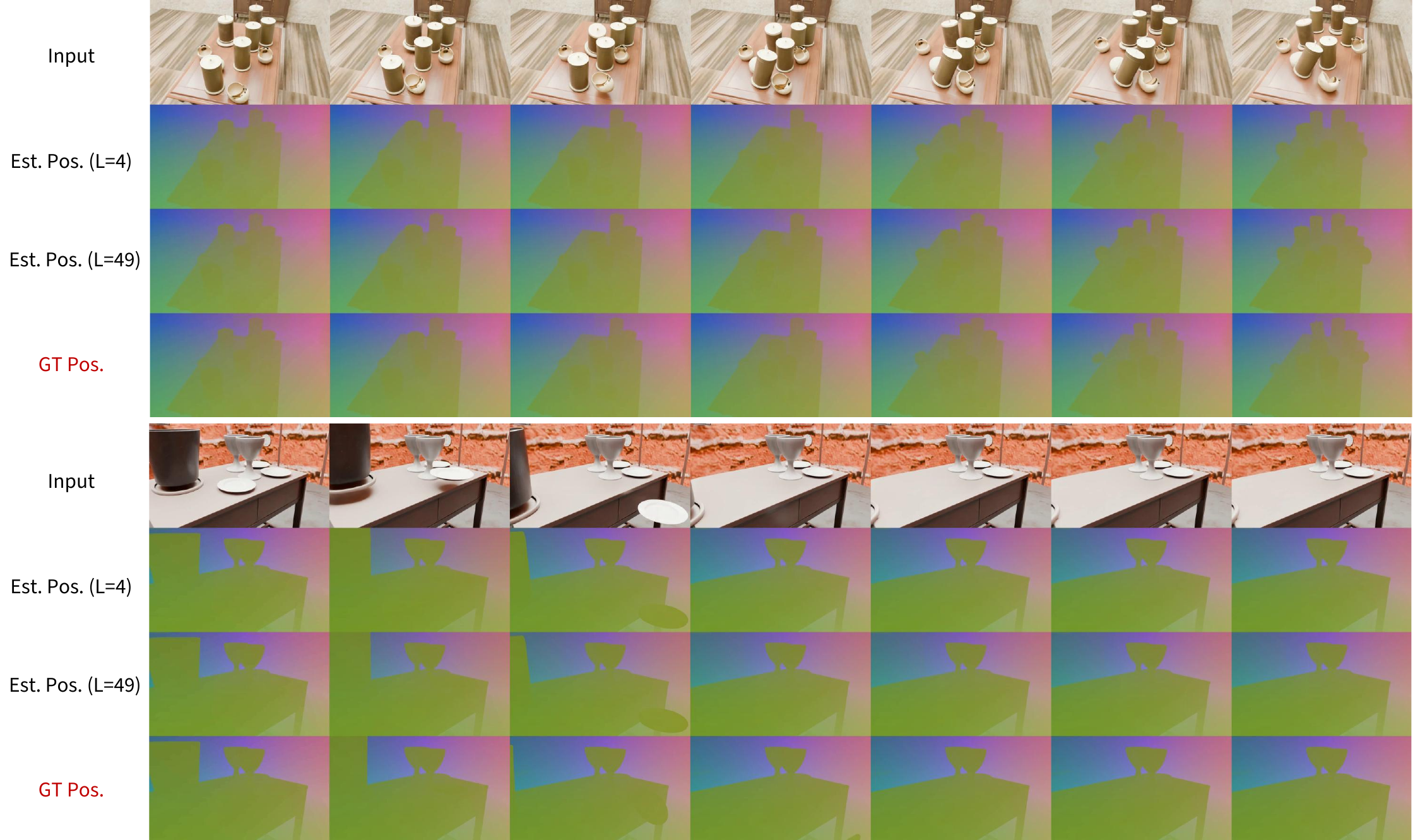}
    \caption{Positional map comparison across window sizes. Each panel shows $7$ uniformly sampled frames. Rows from top to bottom: generated RGB, Depth-Anything-3 positional map with $L{=}4$ (one latent frame), Depth-Anything-3 positional map with $L{=}49$ (full sequence), and ground-truth positional map. The $L{=}4$ and $L{=}49$ results are visually indistinguishable.}
    \label{fig:posmap_window}
\end{figure*}

We further compare the visual quality of the positional map estimated with a small window ($L{=}4$ pixel frames, i.e., one latent frame) against a full-sequence window ($L{=}49$, all frames).
\Cref{fig:posmap_window} shows two representative cases; each panel displays seven uniformly sampled frames, with rows corresponding to the generated RGB, the $L{=}4$ positional map, the $L{=}49$ positional map, and the ground-truth positional map.
Visually, the $L{=}4$ and $L{=}49$ results are nearly indistinguishable, confirming that a minimal window of one latent frame is sufficient for consistent positional-map estimation in our static-camera setting.

\section{Tracking Map: Effect over Long Horizons}
\label{sec:supp_track_long}

In \cref{sec:ablation} we quantify the importance of the tracking map on the long-horizon benchmark by dropping the estimated tracking map from the full model at inference, either entirely or after frame $100$; the evaluation metrics are defined in \cref{sec:supp_lh_metrics}. As \cref{tab:longhorizon_ablation} shows, the response rate drops clearly without the tracking map, especially over long horizons ($90.3\%\rightarrow85.0\%$ for events after frame $100$), while control accuracy stays within noise; withdrawing the map midway ($87.6\%$) sits in between, i.e., a mid-generation tracking failure degrades responsiveness gracefully rather than derailing generation. The tracking map is what keeps late control signals effective, complementing the qualitative zig-zag ablation in \cref{sec:ablation}.

\input{tables/longhorizon_ablation}

\section{Comparison of Autoregressive Training Paradigms}
\label{sec:supp_paradigm}

Teacher-Forcing is known to suffer from exposure bias, so we compare it against Diffusion-Forcing~\cite{chen2024diffusion} and Self-Forcing~\cite{huang2025self} trained from the same Stage-1 model under the same compute budget (best checkpoint each; \cref{tab:paradigm}). Diffusion-Forcing uses the identical architecture and conditions but denoises each frame with independently sampled noise instead of clean teacher context. Self-Forcing distills a $4$-step causal student with distribution matching on its own rollouts; since the online estimators cannot run inside every training rollout, it is trained with the velocity condition only. Teacher-Forcing remains best overall ($5/8$ metrics): Diffusion-Forcing shares the exposure-bias issue yet performs worse across the board, and Self-Forcing removes exposure bias but performs no better---its photometric consistency drops sharply ($75.4$ vs.\ $81.7$), echoing configuration~(a) in \cref{tab:ablation}, which indicates that bidirectional-to-few-step-causal distillation transfers the distribution imperfectly. Teacher-Forcing is therefore the most suitable paradigm for our task, while a distilled few-step student remains attractive for speed (see \cref{sec:runtime}).

\input{tables/paradigm}

\section{Metric Details}
\label{sec:supp_metrics}

We provide full definitions and implementation details for all metrics used in the main text.

\subsection{Object Consistency (ObjCon)}

Per-object DINO ViT-B/16~\cite{caron2021emerging} feature similarity across frames, adapted from VBench++~\cite{huang2025vbench++}.
Dynamic objects are identified via the GT trajectory palette; each object is tracked through the generated video using SAM2~\cite{ravi2024sam}, and tight bounding-box crops (with $8$\,px padding) are extracted per frame.
Only ``interior'' frames are counted: mask area $\geq 200$\,px and no mask pixel within $5$\,px of the image boundary (edge-exit cutoff).

The per-object score is:
\begin{equation}
\text{ObjCon}^{(o)} = 0.4 \cdot \overline{s}_{\mathrm{ref}}^{(o)} + 0.3 \cdot \overline{s}_{\mathrm{consec}}^{(o)} + 0.3 \cdot \min(s_{\mathrm{consec}}^{(o)}),
\end{equation}
where $s_{\mathrm{ref},t} = \cos(\phi(c_t), \phi(c_0))$ and $s_{\mathrm{consec},t} = \cos(\phi(c_t), \phi(c_{t-1}))$, with $\phi$ denoting DINO ViT-B/16 features. The final ObjCon is the mean over all dynamic objects.

\textbf{Modification from VBench:} VBench computes consistency on full frames; we instead crop and mask each object individually, which prevents the static background from dominating the score.

\subsection{Scene Consistency (ScnCon)}

Same formula as ObjCon but computed on full $224{\times}224$ resized frames (no cropping/masking), directly from VBench++~\cite{huang2025vbench++}.

\subsection{Photometric Consistency (PhotoC)}

Forward--backward optical-flow cycle consistency following WorldScore~\cite{duan2025worldscore}.
We compute RAFT-Large~\cite{teed2020raft} forward and backward flow between consecutive frames and measure the average end-point error:
\begin{equation}
\text{AEPE}_{fb}(t) = \frac{1}{|\Omega|}\sum_{\mathbf{p} \in \Omega} \left\| \mathbf{F}_{fw}(\mathbf{p}) + \mathbf{F}_{bw}\!\bigl(\mathbf{p} + \mathbf{F}_{fw}(\mathbf{p})\bigr) \right\|_2,
\end{equation}
where $\Omega$ excludes a $15$-pixel border. The final score is normalized to $[0, 100]$: $\text{PhotoC} = (1 - \text{clamp}(\overline{\text{AEPE}_{fb}} / 1.192, 0, 1)) \times 100$.

\subsection{Trajectory ADE and ADE-Median}

We sample $32$ query points per dynamic object uniformly within the GT mask at frame $0$, then run CoTracker3~\cite{karaev2025cotracker3} on both the GT and generated videos.
Per-object scoring starts from the first frame where the GT object begins moving (mean displacement $> 10$\,px over a $5$-frame lookahead).
\begin{equation}
\text{ADE\_r} = \frac{\sum_o w_o \cdot \bar{e}_o}{\sum_o w_o}, \quad \bar{e}_o = \frac{1}{|M_o|}\sum_{(t,k)\in M_o} \|\hat{\mathbf{x}}^{\text{pred}}_{t,k} - \hat{\mathbf{x}}^{\text{gt}}_{t,k}\|_2,
\end{equation}
where $M_o$ contains GT-visible frame-point pairs within the motion window and $w_o = |M_o|$.
ADE-Median replaces the per-object mean with the median for robustness to outliers.

\subsection{Failure Rate}

Fraction of GT-visible tracked points where the generated video's track is either lost (GT visible but prediction invisible) or deviates by more than $30$\,px:
\begin{equation}
\text{Fail\%} = \frac{\sum_{(t,k) \in M} \mathbb{1}\!\bigl[(\neg v^{\text{pred}}_{t,k}) \lor (\|e_{t,k}\| > 30)\bigr]}{|M|} \times 100.
\end{equation}
The $30$\,px threshold is deliberately strict.

\subsection{FVD}

Complementing the description in \cref{sec:metrics}, we use the $400$-dim logits output of the I3D network as the feature representation.
We sample $16$ frames evenly from each video, resize to $224{\times}224$, and compute the Fréchet distance between the GT and generated feature distributions.
The GT reference set consists of $2500$ videos rendered specifically for this purpose.

\subsection{FVMD}

The motion features of FVMD are extracted with PIPs++ point tracking.
Each $49$-frame video yields $34$ overlapping $16$-frame windows (stride $1$); $400$ points are tracked per window at $256{\times}256$ resolution.
The Fréchet distance is computed between concatenated velocity + acceleration histogram features of GT and generated sets. Over the $64$-video evaluation set this yields $2{,}176$ per-window motion samples for the Fréchet statistics.

\subsection{MLLM Evaluation (SA and PC)}

Following VideoPhy~\cite{bansal2024videophy}, we query GPT-4o with the input image (frame $0$), $8$ evenly spaced generated frames, and a JSON specification of the intended velocity-increment events.
GPT-4o rates each video on two axes (1--5 Likert scale):
\textbf{Semantic Adherence (SA)}: how well the generated content and motion match the scene description and velocity directions.
\textbf{Physical Commonsense (PC)}: whether the resulting object motion is intuitively and physically plausible.
We use temperature $0.3$ and max $512$ tokens.
The full system prompt is shown below.

\begin{figure}[h]
\small
\begin{tcolorbox}[colback=gray!5, colframe=gray!50, title=MLLM System Prompt for SA/PC Evaluation, fonttitle=\bfseries\small]
You are evaluating a physics-grounded image-to-video generation model.

You will receive:
\begin{enumerate}[leftmargin=*, nosep]
\item An input image showing a static indoor scene with rigid-body objects on a surface.
\item A kick specification (JSON) describing instantaneous velocity impulses applied to objects at specific frames --- this is the intended physical interaction.
\item A sequence of evenly-spaced frames from the generated video.
\end{enumerate}

\textbf{Kick Specification:} Each kick has: \texttt{frame} (0--48), \texttt{object\_id}, \texttt{type} (``A'' = horizontal only; ``B'' = includes upward component), \texttt{v\_cam} = $[v_x, v_y, v_z]$ in camera coordinates ($v_x{>}0$: right; $v_y{>}0$: up; $v_z{>}0$: toward camera), and \texttt{v\_scale} (0.5--1.0, higher = faster).

\textbf{Evaluation Criteria} (1--5 Likert):
\begin{enumerate}[leftmargin=*, nosep]
\item \textbf{Semantic Adherence (SA):} Does the video start from the input image with recognizable objects and preserved scene layout?
\item \textbf{Physical Commonsense (PC):} Does the object motion follow physically plausible dynamics given the applied kicks? Consider: correct direction, realistic sliding/tumbling/bouncing, friction-based deceleration, plausible collisions, and gravity effects.
\end{enumerate}

\textbf{Output:} Return exactly one line: \texttt{SA=X, PC=Y} where X, Y $\in \{1,2,3,4,5\}$.
\end{tcolorbox}
\end{figure}

\subsection{Human Preference}

We conduct a user study with $25$ evaluators on $20$ in-the-wild test cases (our method + $4$ baselines = $5$ videos per case).
Videos are anonymized and randomly shuffled via a Latin-square design.
Each evaluator selects the best video for three criteria independently:
\textbf{Physical Plausibility (Phys.)}: most physically realistic motion;
\textbf{Motion Accuracy (Motn.)}: best match to the specified velocity directions and affected objects;
\textbf{Visual Quality (Vis.)}: best overall visual and temporal quality.
Results are reported as win rate (\%) per method per criterion; the uniform baseline is $20\%$.
The full guidance shown to evaluators is reproduced below.

\begin{figure}[h]
\small
\begin{tcolorbox}[colback=gray!5, colframe=gray!50, title=Human Preference Study Guidance, fonttitle=\bfseries\small]
You will be presented with 20 questions, each involving a short video of a tabletop scene in which everyday rigid objects are pushed by unseen forces---sliding, tumbling, bouncing, and colliding as solid bodies.

For each question, you will see an input image alongside a control video that visualizes the applied forces. In the control video, red arrows indicate forces acting within the horizontal plane, while blue arrows indicate forces that contain an upward component against gravity. Forces may be applied to multiple objects and may appear at any intermediate frame, indicating the moment at which the force begins to act.

Below the input image and control video are five generated videos (A--E). Please evaluate them along three criteria:

\begin{enumerate}[leftmargin=*, nosep]
\item \textbf{Physical Plausibility:} Select the video in which the objects move, collide, and come to rest in the most physically realistic manner---obeying gravity, conservation of momentum, and rigid-body contact dynamics.
\item \textbf{Motion Accuracy:} Select the video whose object motion best matches the forces depicted in the control video---correct direction, affected objects, and timing.
\item \textbf{Visual Quality:} Select the video with the best overall visual and temporal quality---sharpness, consistency, and absence of artifacts.
\end{enumerate}

For each criterion, click A / B / C / D / E to select your preferred video.
\end{tcolorbox}
\end{figure}

\subsection{Metrics for the Long-Horizon Benchmark}
\label{sec:supp_lh_metrics}

The long-horizon benchmark (\cref{sec:longvideo}) contains $5$ multi-object scenes of $301$ frames; every object receives a velocity increment every $24$ frames with the direction rotating by $45^{\circ}$ per event ($231$ events in total), and all metrics are computed per $100$-frame segment. Let $\mathbf{c}_{o}(t)$ denote the SAM2-tracked centroid of object $o$ and $\mathbf{v}_{o}(t)=\mathbf{c}_{o}(t{+}1)-\mathbf{c}_{o}(t)$ its per-frame velocity.

\topic{Average Consistency.}
The mean of the scene, object, and photometric consistency metrics defined above, computed over the frames of each segment, with the reference frame kept at frame $0$ of the full video.

\topic{Successful Respond.}
For a control event $k=(f_k, o_k, \mathbf{v}^{\mathrm{cam}}_k)$, the mean-velocity change over a window $W{=}5$ is
\begin{equation}
\Delta\bar{\mathbf{v}}_k \;=\; \tfrac{1}{W}\textstyle\sum_{t=f_k}^{f_k+W-1}\mathbf{v}_{o_k}(t) \;-\; \tfrac{1}{W}\textstyle\sum_{t=f_k-W}^{f_k-1}\mathbf{v}_{o_k}(t),
\end{equation}
where the second term is $\mathbf{0}$ for $f_k{=}0$. The target object counts as responding if $\lVert\Delta\bar{\mathbf{v}}_k\rVert \ge 0.3$\,px/frame, or if its masked region shows an appearance discontinuity---the mean consecutive-frame SSIM before the event exceeds the post-event minimum by at least $0.05$---which catches touching objects whose centroids barely move. Events whose object can no longer be tracked by SAM2 are excluded from the denominator; Successful Respond is the fraction of the remaining (verifiable) events with a response.

\topic{Control Accuracy.}
The commanded direction is the image-plane projection of the event velocity (the image $y$-axis points down), $\mathbf{d}_k=\big(v^{\mathrm{cam}}_{x,k},\,-v^{\mathrm{cam}}_{y,k}\big)$, and $\cos\theta_k=\Delta\bar{\mathbf{v}}_k\cdot\mathbf{d}_k\,/\,\big(\lVert\Delta\bar{\mathbf{v}}_k\rVert\,\lVert\mathbf{d}_k\rVert\big)$. For a segment $\mathcal{S}$, let $\mathcal{R}_{\mathcal{S}}$ be the responded events in $\mathcal{S}$ whose direction is measurable ($\lVert\Delta\bar{\mathbf{v}}_k\rVert\ge 0.3$\,px/frame and $\lVert\mathbf{d}_k\rVert\ge 0.15\,\lVert\mathbf{v}^{\mathrm{cam}}_k\rVert$); then
\begin{equation}
\text{Control Accuracy}(\mathcal{S}) \;=\; \frac{100}{|\mathcal{R}_{\mathcal{S}}|}\sum_{k\in\mathcal{R}_{\mathcal{S}}}\frac{1+\cos\theta_k}{2},
\end{equation}
i.e., $100$ means the responded motion is perfectly aligned with the command, $50$ orthogonal, and $0$ opposite.

\section{In-the-Wild Evaluation Details}
\label{sec:supp_wild}

The in-the-wild evaluation set consists of $20$ input images sourced from real-world photographs and high-quality text-to-image generations.
For each image, the user specifies a set of target objects (via point-click segmentation) and a sequence of velocity-increment events at chosen frames, following the same interface as the synthetic benchmark.
The control signals are randomly generated under the same rules as the training data (kick-A/B types, bounded by $V_{\max}$) to avoid cherry-picking.

The MLLM evaluation and human preference study are described in detail in the metric sections above.
The user study collected responses from $25$ evaluators, each rating all $20$ cases.

\section{Additional In-the-Wild Results}
\label{sec:supp_more_results}

\begin{figure*}[t]
    \centering
    \includegraphics[width=\textwidth]{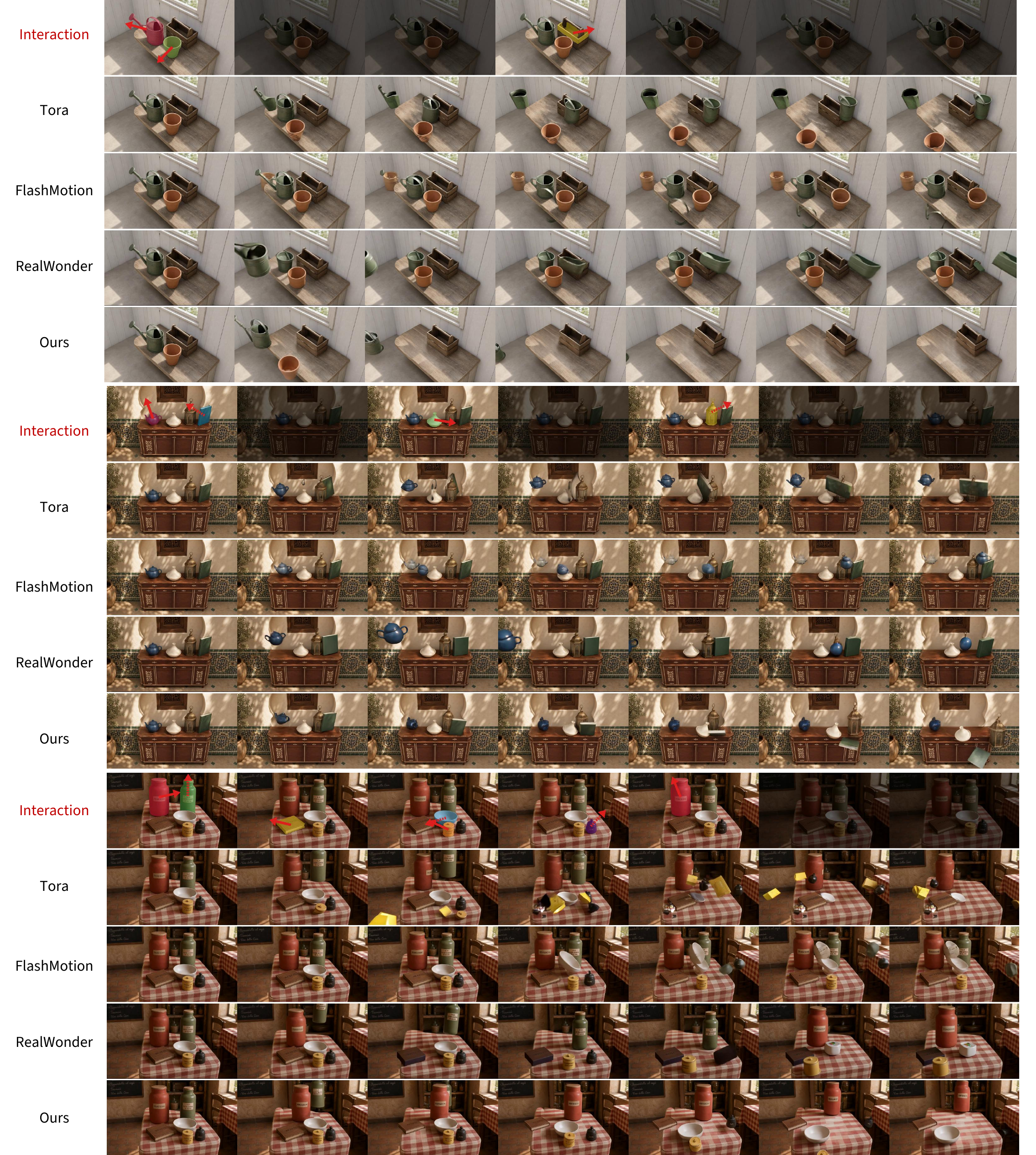}
    \caption{Additional in-the-wild results. Each row shows an input image with user-specified velocity-increment interactions, followed by representative frames from the \name generation.}
    \label{fig:more_results}
\end{figure*}

\Cref{fig:more_results} presents additional qualitative results on in-the-wild input images, demonstrating that \name generalizes to diverse real-world scenes with physically plausible multi-object dynamics.

%% file: tables/longhorizon_ablation.tex
\begin{table}[t]
    \caption{Tracking-map ablation on the long-horizon benchmark: the estimated tracking map is kept (Default), dropped after frame $100$, or dropped throughout, at inference; 1--100 / 101+ denote frame ranges.}
    \label{tab:longhorizon_ablation}
    \centering\small
    \setlength{\tabcolsep}{4pt}
    \begin{tabular}{l|cc|cc}
        \toprule
& \multicolumn{2}{c|}{Successful Respond\,$\uparrow$} & \multicolumn{2}{c}{Control Accuracy\,$\uparrow$} \\
Tracking map & 1--100 & 101+ & 1--100 & 101+ \\
        \midrule
Default            & 100.0\% & 90.3\% & 95.2\% & 81.2\% \\
Dropped after 100  & 100.0\% & 87.6\% & 95.3\% & 82.0\% \\
Dropped entirely   & 97.4\%  & 85.0\% & 96.4\% & 80.1\% \\
        \bottomrule
    \end{tabular}
\end{table}

%% file: tables/paradigm.tex
\begin{table*}[!tp]
    \caption{Comparison of autoregressive training paradigms on test set~(ii). All variants start from the same Stage-1 model and are trained under the same compute budget (best checkpoint each); Self-Forcing is trained with the velocity condition only. Column abbreviations follow \cref{tab:main_results}.}
    \label{tab:paradigm} 
    \centering \small
    \setlength{\tabcolsep}{4pt}
    \renewcommand{\arraystretch}{1.1}
    \begin{tabular}{l|cc|ccc|ccc}
        \toprule
& FVD\,$\downarrow$ & FVMD\,$\downarrow$ & Traj-ADE\,$\downarrow$ & Traj-ADE-M\,$\downarrow$ & Failure\,$\downarrow$ & Scene Cons.\,$\uparrow$ & Obj. Cons.\,$\uparrow$ & Photo. Cons.\,$\uparrow$ \\
        \midrule
Diffusion-Forcing               & 397.9 & 799.7 & 41.05 & 31.76 & 45.95 & 96.13 & 83.83 & 80.78 \\
Self-Forcing (velocity only)    & \textbf{387.0} & 937.3 & \textbf{40.19} & \textbf{31.32} & 46.55 & 95.94 & 85.16 & 75.40 \\
Teacher-Forcing (Ours)          & 413.7 & \textbf{787.0} & 40.24 & 32.00 & \textbf{43.15} & \textbf{96.57} & \textbf{85.29} & \textbf{81.72} \\
        \bottomrule
    \end{tabular} 
\end{table*}